\documentclass[journal]{IEEEtran}
\usepackage{amsmath,amssymb,booktabs,graphicx,microtype,xcolor,array,url,cite,placeins}
\usepackage{bibunits}
\graphicspath{{figures/}}
\newcommand{\method}{EBV-CAN}
\title{EBV-CAN: Vision--Language Grounding for Contact-Aided Endoscope Navigation}
\author{Chi~Kit~Ng$^{1,*}$, Shuchang~Liu$^{1,*}$, Jinsong~Lin$^{1,*}$,\\
Yidong~Zhang$^{1}$, Xizi~Wang$^{1}$, Ho~Yin~Chim$^{2}$, Hongliang~Ren$^{1,\dagger}$%
\thanks{$^{1}$The Chinese University of Hong Kong. $^{2}$National University of Singapore. $^{3}$Stanford University.}%
\thanks{$^{*}$Chi Kit Ng, Shuchang Liu, and Jinsong Lin contributed equally. $^{\dagger}$Corresponding author: Hongliang Ren.}}
\begin{document}
\maketitle
\begin{bibunit}[IEEEtran]
\begin{abstract}
Flexible endoscopes exploit structural compliance to traverse narrow lumens, pipes, and cavities, where navigation involves distributed environmental contact. This compliance makes tip motion depend on the evolving support configuration. We present external bird's-eye-view contact-aided navigation (EBV-CAN), a vision-language policy that combines one initial global static bird's-eye image with evolving endoscopic observations and motor state. Three supervised semantic queries locate the tip, next waypoint, and instructed target on the cached global image; their residual representations predict bending and insertion commands. We introduce the Vision-Language Contact-Aided (VLCA) dataset, linking 517 demonstrations and 75k action samples to 958k spatial entries and 93k remaining-route records. On 50 held-out episodes, tip-coordinate and waypoint localization reach 97.5\% and 94.8\%. Swapping the target instruction under identical observations changes target attention and motor output across 17 dual-target episodes. Across 1,920 physical trials of eight policies trained under a shared protocol, EBV-CAN achieves 183/240 successes (76.25\%) under automatic blue-area scoring. Gated recurrent unit behavior cloning (GRU-BC) achieves the lowest matched-task offline mean absolute error (MAE), 0.056, and 1/240 physical successes, establishing distinct rankings for action imitation and task completion. Successful multi-guide executions and representative failures connect spatial progress, steering transitions, and visibility to physical navigation.

\end{abstract}

\section{Introduction}
\IEEEPARstart{F}{lexible} endoscopes and continuum robots combine a slender body with distributed compliance, enabling access through confined and curved spaces \cite{burgner2015continuum,rus2015design,kim2022advancement}. Navigation couples perception, robot mechanics, and sequential decisions: the controller selects a route, advances the shaft, and adjusts the distal view as the robot interacts with its surroundings. Reviews of medical robotics and autonomous intraluminal navigation identify this coordination as a central component of increasing robotic autonomy \cite{dupont2021decade,pore2023autonomous}. For a flexible endoscope with a short actively bending tip and a long passive shaft, the control problem extends along the body. Insertion changes the loaded shaft configuration, while friction and external support change the relation between motor commands and tip displacement \cite{webster2010design,bajo2011kinematics}.

\begin{figure}[t]
\centering
\includegraphics[width=\columnwidth]{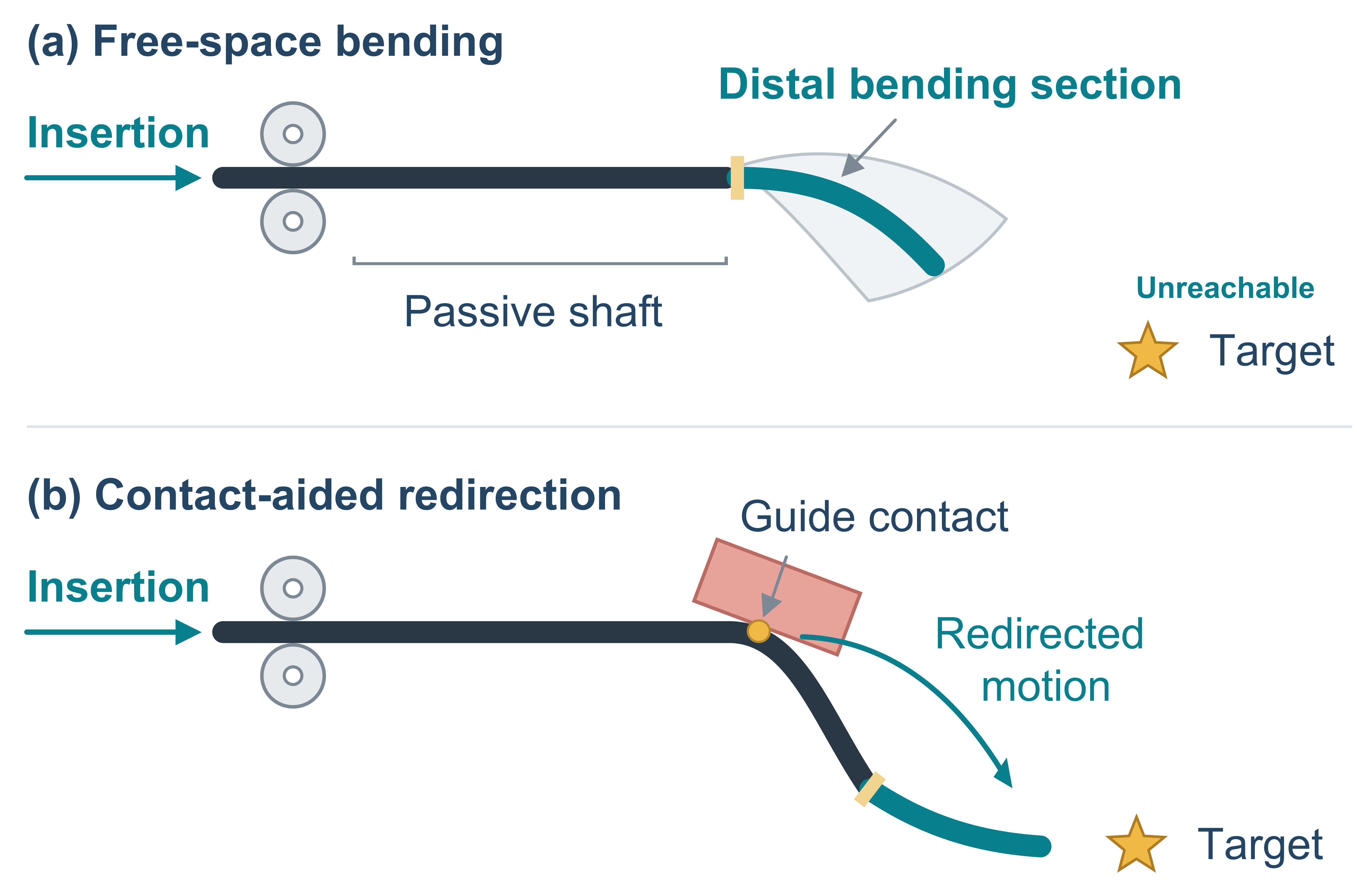}
\caption{Contact-aided navigation mechanism. A guide supports the long passive shaft, allowing insertion and distal bending to redirect the endoscope toward a target beyond the illustrated free-space bending envelope. The envelope is a qualitative schematic.}
\label{fig:concept}
\end{figure}

Environmental support provides a mechanism for contact-aided navigation (CAN). A guide constrains part of the passive shaft, creating a support location around which further insertion and distal bending redirect the tip (Fig.~\ref{fig:concept}). Rao \emph{et al.} exploit environmental contact to produce multiple curvatures in a single-segment tendon-driven robot through motion planning \cite{rao2024towards}. Ng \emph{et al.} demonstrate contact-aided endoscope navigation with force-informed reinforcement learning in a deformable stomach simulation \cite{ng2025contact}. These results motivate learning the timing of guide approach, supported bending, and subsequent advancement from physical demonstrations. In this setting, a successful motor command depends on both the destination and the robot's progress through a sequence of interactions.

Visual feedback offers complementary evidence at two spatial scales. Endoscopic images resolve nearby surfaces and apparent motion; a global view identifies the arrangement of guides and candidate targets. Learned endoscopic controllers already use monocular depth, virtual eye-in-hand observations, and bronchoscopic images for navigation \cite{lu2023autonomous,lai2023sim,zhang2024ai}. Here, the central question is how evolving local evidence updates route progress in a global scene observed once. Near a guide, the endoscopic view (EV) is dominated by nearby geometry, while the instructed destination remains a global property. The controller therefore needs a persistent spatial reference together with a changing estimate of its tip and next waypoint.

Language supplies the target identity within this spatial problem. Language-conditioned imitation and vision-language-action (VLA) models provide mechanisms for mapping task descriptions to behavior \cite{jang2022bc,brohan2023rt,kim2024openvla}. For contact-aided navigation, the practical question is specific: under identical visual observations, does changing the selected target change the spatial prediction and the motor command? Answering this question benefits from annotations that connect an instruction to an object, a route, and a sequence of physical actions. Dense spatial labels also make route progress directly evaluable alongside action imitation.

Evaluation also has a sequential dimension. Offline action mean absolute error (MAE) measures agreement with expert actions at recorded states, whereas physical execution generates subsequent observations through the policy's own actions. This distinction is central to sequential imitation learning \cite{ross2011reduction}. Contact-aided navigation makes it concrete: a small steering change alters the shaft's approach to a guide, the support configuration, and the next visible scene. We therefore evaluate action fidelity, spatial progress, and physical endpoints as complementary properties of a navigation policy.

We study this problem on a planar physical platform with two degrees of freedom (DoFs): bending and insertion. The policy receives one initial external bird's-eye-view (EBV) image, followed by EV images and causal motor state. Continuous feedback comes from the EV and motor channels; the initial EBV provides fixed global context. This observation contract isolates local-to-global estimation under stable scene registration. Contact-related supervision describes visible shaft support and image-space route progress, with visual and motor observations carrying the interaction evidence.

Our external bird's-eye-view contact-aided navigation (EBV-CAN) policy reads frozen initial-EBV features produced by self-distillation with no labels, version 2 (DINOv2), \cite{oquab2023dinov2} using tip, waypoint, and target queries conditioned on local observations and language. Spatial supervision connects those queries to route progress, while their residual representations drive bending and insertion. This couples interpretable spatial outputs to feedback control.

The accompanying \emph{Vision-Language Contact-Aided (VLCA) dataset} links physical demonstrations to object boxes, tip coordinates, canonical guide identities, ordered episode plans, and frame-dependent remaining routes. This combination provides supervision at the observation, spatial-progress, and motor-action levels. Our contributions are:
\begin{itemize}
\item A spatially grounded vision-language control formulation for contact-aided navigation: evolving endoscopic evidence updates tip and route progress on one initial global static image, and language-conditioned residual queries connect these estimates to motor commands.
\item VLCA, a physical navigation dataset that connects dense scene geometry to sequential control through canonical guide identities, ordered plans, and remaining-route labels. Its 517 demonstrations, 74,693 action samples, and 957,616 spatial entries support joint study of localization, route transitions, and action prediction.
\item Physical evidence that action imitation and contact-aided task completion yield distinct policy rankings. Under a shared training protocol and automatic endpoint criterion, EBV-CAN achieves 76.25\% success over 240 trials, exceeding the strongest baseline by 60.0 percentage points; spatial and instruction interventions identify how local feedback and target language affect its predictions.
\end{itemize}

\section{Related Work}
\subsection{Contact-Aided Continuum-Robot Motion}
Contact-aware continuum control addresses how distributed constraints alter robot shape and transmitted motion. Bajo and Simaan localize contacts through kinematic information \cite{bajo2011kinematics}, while Goldman \emph{et al.} use actuation-force sensing for compliant motion \cite{goldman2014compliant}. Embedded sensing and recurrent models connect soft-body deformation to state estimates \cite{thuruthel2019soft}; expected-versus-observed deformation also supports simultaneous proprioception and contact detection \cite{wang2024sensing}. At the mechanism level, contact-aided compliant structures produce anisotropic bending shapes and support anatomy-specific design \cite{ai2021multi,ai2023anatomy}.

At the navigation level, Rao \emph{et al.} explicitly search for obstacle-supported motions of a tendon-driven continuum robot \cite{rao2024towards}. Ng \emph{et al.} learn a force-informed contact-aided endoscope policy in finite-element stomach simulation \cite{ng2025contact}. These approaches establish complementary roles for contact modeling, sensing, and learning. Our formulation contributes a demonstration-based visual control setting: the initial scene defines guide and target locations, and local observations condition spatial predictions and motor output throughout physical execution.

\subsection{Visual Feedback for Endoscopic Navigation}
Visual navigation connects endoscopic observations to steering and advancement. Online nonparametric learning adapts control of a soft continuum robot \cite{lee2017nonparametric}; depth-guided navigation combines monocular perception with three-dimensional shape planning \cite{lu2023autonomous}; virtual eye-in-hand training transfers learned navigation strategies to a physical soft robot \cite{lai2023sim}. Zhang \emph{et al.} integrate bronchoscopic images and human commands in a learned shared-control co-pilot \cite{zhang2024ai}. Across these settings, visual feedback provides task-relevant state information during motion.

Global imaging supplies a complementary route reference. Barnes \emph{et al.} combine a static goal feature map with live simulated fluoroscopic images for goal-conditioned soft-robot navigation and evaluate new vascular geometries \cite{barnes2025toward}. Our observation contract pairs an initial external image with ongoing tip-camera observations on a planar board. Spatial supervision connects the two views through tip, waypoint, and target labels, and the episode-disjoint protocol evaluates new demonstrations from the recorded layout distribution. Each platform's success measure is interpreted within its own imaging and validation protocol.

\subsection{Language-Conditioned Learning and Spatial Supervision}
Visuomotor imitation learns controls from demonstrations whose collection, temporal structure, and evaluation protocol influence performance \cite{mandlekar2021matters}. Gated recurrent unit behavior cloning (GRU-BC) supplies a recurrent baseline. The action chunking transformer (ACT) predicts action sequences \cite{zhao2023learning}, Diffusion Policy models actions by conditional denoising \cite{chi2025diffusion}, and multi-view representations integrate complementary cameras \cite{seo2023multi}. Our model uses a fixed EBV reference and a changing EV history, with spatial labels supplying explicit intermediate objectives for route progress.

Language-conditioned imitation and VLA policies associate instructions with robot behavior \cite{jang2022bc,brohan2023rt,kim2024openvla,team2024octo}. Qwen3 vision-language (Qwen3-VL), Generalist Robot 00 Technology (GR00T), and the StarVLA framework support the larger-backbone configurations evaluated here \cite{bai2025qwen3,bjorck2025gr00t,community2026starvla}. We evaluate frozen backbones and low-rank adaptation (LoRA) \cite{hu2021lora}. \method{} uses a compact text encoder and two target templates to study semantic conditioning under matched observations. Because interpreting attention as an action explanation requires additional evidence \cite{jain2019attention}, we evaluate target-query changes together with motor-output changes and visual-input interventions. VLCA supplies the linked spatial and route-progress annotations needed to examine these behaviors quantitatively.

\section{The VLCA Dataset}
\label{sec:data}
\subsection{Platform, Observations, and Task Design}
\begin{figure*}[!t]
\centering
\includegraphics[width=\textwidth]{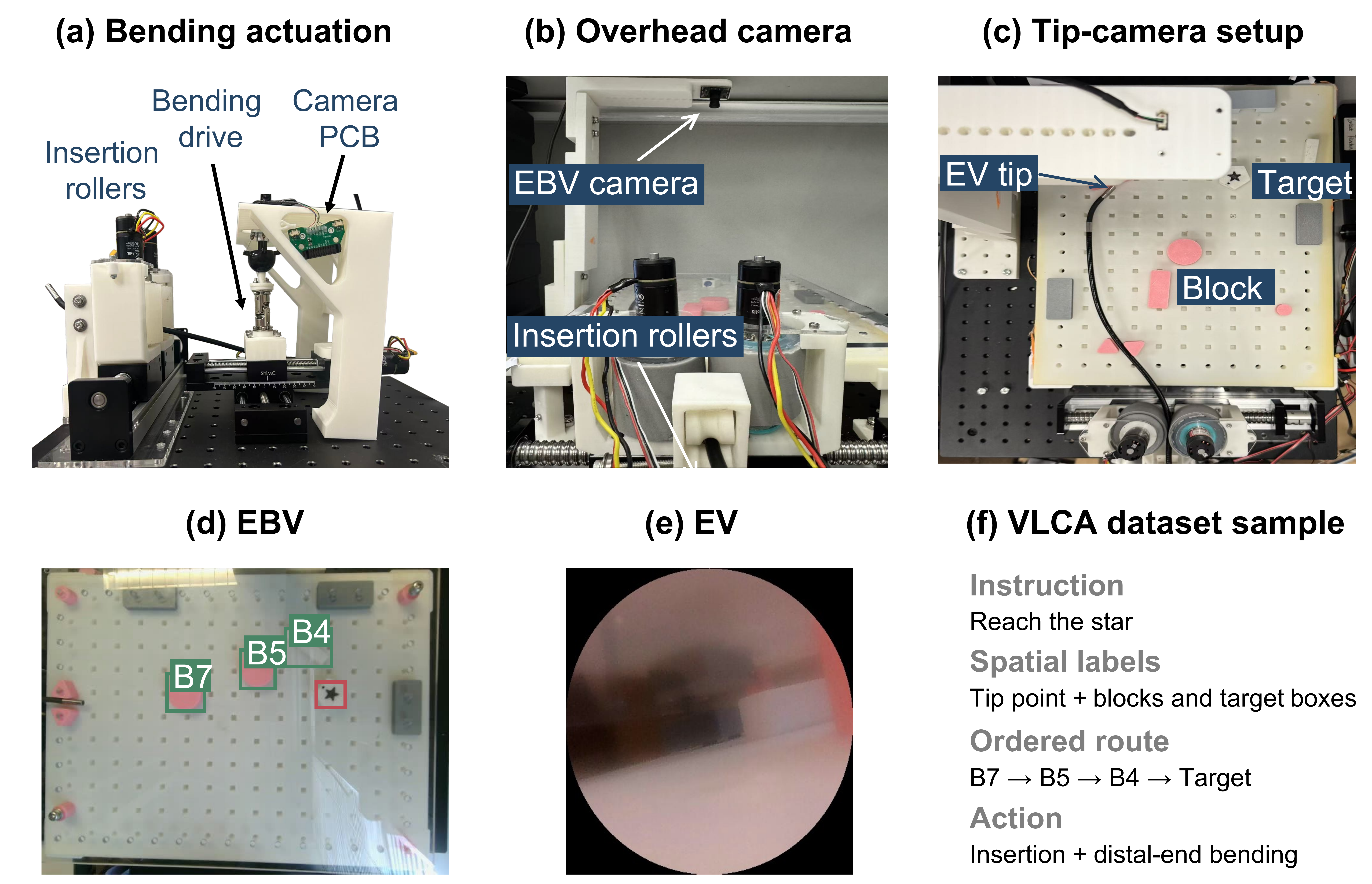}
\caption{Physical endoscope platform. A motorized bending knob controls the distal segment, and counter-rotating rollers control insertion. The overhead camera records the global scene; the illuminated tip camera supplies the EV. PCB denotes the printed circuit board.}
\label{fig:platform}
\end{figure*}

The platform motorizes an industrial flexible endoscope with two command channels: distal planar bending through a motorized knob, and shaft insertion/retraction through counter-rotating friction rollers. The proximal shaft remains passive. A fixed overhead camera observes a planar board with guides and target objects, and a tip-mounted illuminated camera supplies the EV stream. Demonstrations are collected through Xbox-controller teleoperation. Figure~\ref{fig:platform} shows the apparatus and its linked supervision.

\begin{figure*}[!t]
\centering
\includegraphics[width=\textwidth]{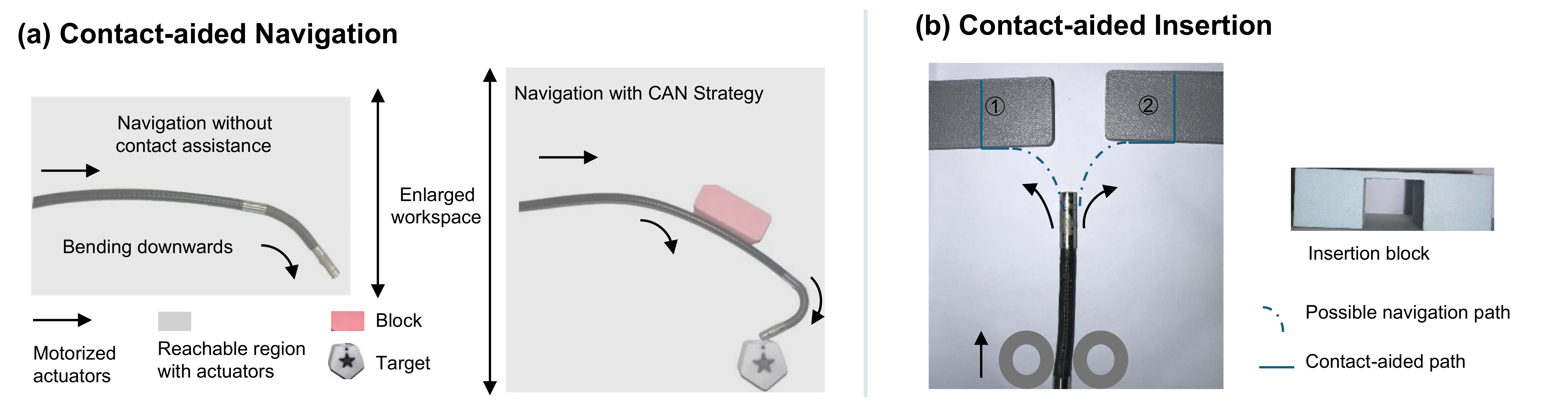}
\caption{Complementary uses of environmental support. (a) Contact-aided navigation redirects the bending section toward the target. (b) Contact-aided insertion constrains shaft progression through a guide. The workspace envelopes and paths illustrate the mechanisms qualitatively.}
\label{fig:contact-modes}
\end{figure*}

Figure~\ref{fig:contact-modes} distinguishes supported redirection from constrained insertion. Both occur within the same two control DoFs, coupling distal bending to the position of the passive shaft.

Nine offline task levels vary guide use, insertion-dependent traversal, and target selection (Table~\ref{tab:dataset}). L0 has a direct-to-target plan; L1, L2, and L3 contain one, two, and three planned guides. Corresponding .5 variants emphasize insertion-dependent traversal, with both command channels available throughout. L4 and L4.5 contain both star and circle targets. L4 plans contain zero to two guides; L4.5 plans contain one. Each level identifies a task family through its route, target, and insertion requirements. The plan specifies the relevant subset of visible blocks.

Each episode has one target-conditioned instruction. Two recorded templates select star or circle and include a shared directive to use guides. The episode-specific guide sequence is stored in the ordered route annotation. This separation supports a controlled study of target-language conditioning and visually grounded route progress.

\begin{table}[t]
\caption{VLCA policy corpus. Guides are those in the annotated plan; frames are retained action samples. I indicates an insertion-dependent task variant.}
\label{tab:dataset}\centering\small
\setlength{\tabcolsep}{4pt}
\begin{tabular}{lclrr}\toprule
Level & Guides & Target / variant & Episodes & Frames\\\midrule
L0 & 0 & Star & 81 & 10,080\\
L1 & 1 & Star & 44 & 7,613\\
L1.5 & 1 & Star / I & 111 & 15,538\\
L2 & 2 & Star & 50 & 9,837\\
L2.5 & 2 & Star / I & 19 & 2,889\\
L3 & 3 & Star & 15 & 3,295\\
L3.5 & 3 & Star / I & 14 & 2,349\\
L4 & 0--2 & Star or circle & 93 & 11,715\\
L4.5 & 1 & Star or circle / I & 90 & 11,377\\\midrule
Total & & & 517 & 74,693\\\bottomrule
\end{tabular}
\end{table}

\subsection{Recording and Policy Samples}
The deposited collection contains 518 recordings. Filtering for a nonempty action export retains 517 episodes and 74,693 nonzero-action rows; L4/0066 is the excluded recording. All retained rows have tip and next-waypoint labels. Each row links an EV frame, motor positions, current action, previous action, instruction, and spatial supervision to the episode's initial EBV. Later EBV images support annotation and visualization; deployment uses the initial EBV throughout.

Both cameras record at $640\times480$ pixels, nominally at 20~Hz. The EV is postprocessed by center cropping to $480\times480$ and applying an inscribed circular mask. Training and physical inference use the same circular-view preprocessing, followed by each encoder's specified resizing. Filtering and acquisition gaps make the retained sequence nonuniform: the median adjacent interval is 50.2~ms, and 2.62\% of adjacent retained intervals exceed 75~ms. An eight-observation window spans seven intervals, with median duration 0.351~s and 95th percentile 0.604~s. Model indexing follows retained observations, preserving their recorded temporal spacing.

Insertion and bending commands are normalized by controller scales 20 and 150 and reported in controller units. The state comprises clipped normalized motor positions and the preceding normalized action, initialized to zero at episode start. The input follows causal observation order, while current and future expert commands supply the supervised output targets.

\subsection{Spatial Annotation and Route Progress}
The annotation export contains 94,173 frame-level EBV records and 957,616 spatial entries: 770,788 guide/object boxes, 93,415 target boxes, and 93,413 robot-tip entries. Bounding boxes use normalized image coordinates $(x,y,w,h)$, and tip locations use normalized $(x,y)$. The separate 93,413-point coordinate field represents the same tip annotations. Table~\ref{tab:annotations} distinguishes the spatial inventory from its route and control links.

\begin{table}[t]
\caption{Linked VLCA annotation inventory. Counts refer to stored entities; tip points and tip entries share underlying annotations.}
\label{tab:annotations}\centering\small
\setlength{\tabcolsep}{4pt}
\begin{tabular}{lrl}\toprule
Entity & Count & Supervision role\\\midrule
Guide/object boxes & 770,788 & Scene geometry\\
Target boxes & 93,415 & Goal grounding\\
Robot-tip entries & 93,413 & Localization\\
Canonical reference boxes & 4,041 & Guide identity\\
Ordered episode plans & 518 & Route order\\
Remaining-route records & 93,085 & Next waypoint\\
Recorded plan updates & 495 & Route progression\\
Aligned action samples & 74,693 & Two-DoFs control\\\bottomrule
\end{tabular}
\end{table}

The annotation interface supports detector proposals, manual box and tip editing, canonical guide assignment, route ordering, and label propagation. A census identifies 4,041 canonical reference boxes and 518 ordered plans. Canonical guide identifiers (IDs) link spatial geometry to each plan, which ends at the target. The 93,085 per-frame remaining-plan records identify the next waypoint: 25,709 point to a guide and 67,376 point to the target. Across these records, 495 changes in the remaining list encode route progression. These links turn a collection of object detections into supervision for sequential navigation.

The progress rule removes a reached guide when the annotated tip approaches its box edge within a configured threshold, retaining the final target. Most recordings use $1/32$ times the maximum image dimension (20 pixels at $640\times480$). The resulting labels describe image-space route progress. Supplementary Information (SI), Sec.~S2, details the annotation classes, thresholds, workflow, and structured-language exports.

\subsection{Partitions and Dataset Contribution}
Episodes are split into 417 training, 50 validation, and 50 test episodes, containing 60,800, 6,614, and 7,279 action samples, respectively. The split is stratified by task level and target identity with seed 42. Whole episodes remain within one partition, while layouts are shared across the recorded distribution. The test therefore measures performance on held-out demonstrations of these task families.

VLCA contributes a linked physical benchmark for three questions: action imitation, spatial estimation of the tip and remaining plan, and output dependence on a target instruction under fixed observations. Dense global annotations supervise learning of a local-to-global estimate, while execution uses one initial overhead image. The 495 route updates additionally support event-centered evaluation around waypoint changes.

\section{Policy Formulation}
\label{sec:method}
\subsection{Static Map and Dynamic Navigation Feature}
\begin{figure*}[!t]
\centering
\includegraphics[width=.99\textwidth]{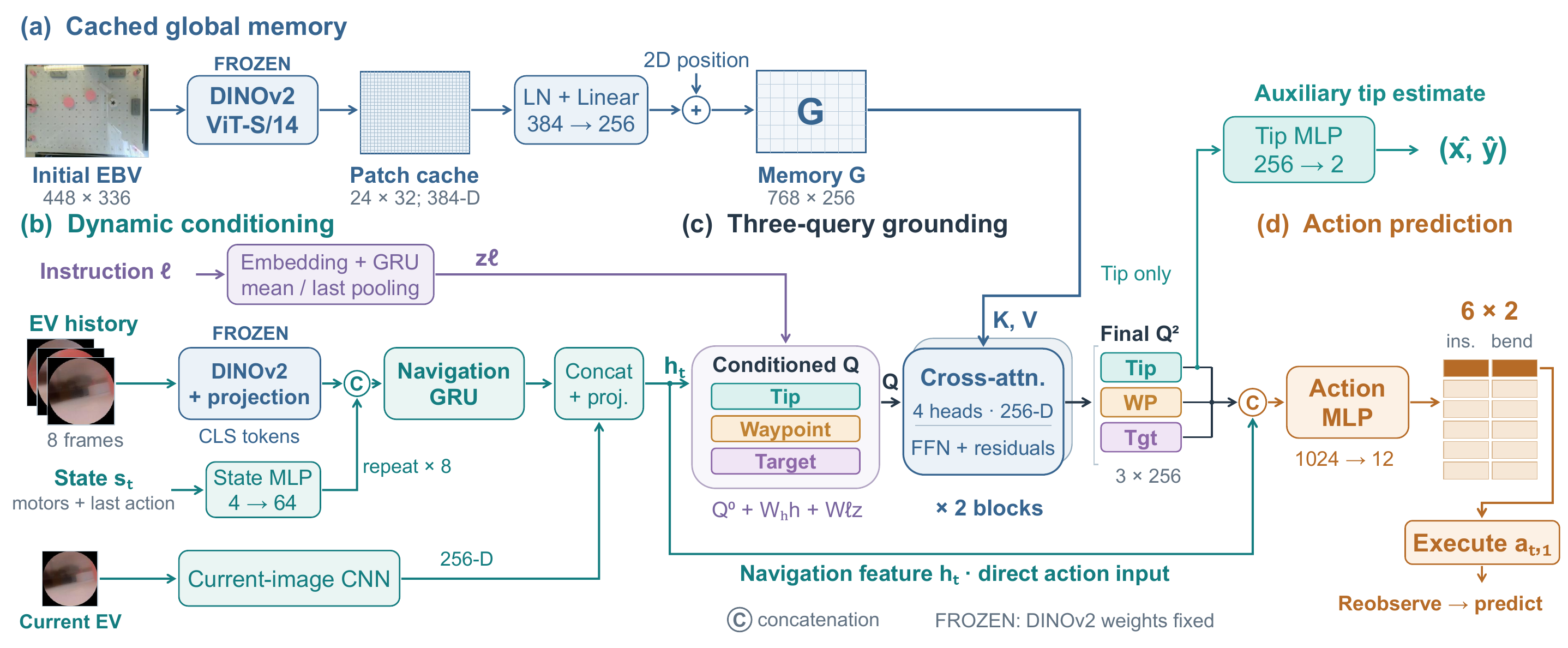}
\caption{Implemented \method{} architecture. (a) Initial-EBV patch features are cached, projected, and combined with fixed two-dimensional positional encodings. (b) EV history, the repeated current-state embedding, and current-image CNN features produce $h_t$; the text encoder produces $z_l$. (c) Both condition all three learned queries. Two residual cross-attention blocks read the fixed memory. (d) The action head concatenates $h_t$ with all three final query tokens and predicts six two-motor commands; only the first is executed. The auxiliary coordinate head uses only the tip token. C denotes concatenation; CLS denotes the class token. Q, K, and V denote query, key, and value; WP and Tgt denote waypoint and target. Input thumbnails are representative.}
\label{fig:architecture}
\end{figure*}

Let $G$ denote the initial EBV, $E_{t-7:t}$ the eight most recent retained EV observations, $s_t$ the causal motor state, and $l$ the instruction. The policy predicts
\begin{equation}
\hat A_t=\pi_\theta(G,E_{t-7:t},s_t,l)\in[-1,1]^{6\times2}.
\end{equation}
Only $\hat A_{t,1}$ is applied before a new prediction. Visual history and motor state provide the dynamic observations. Figure~\ref{fig:architecture} gives the implemented computation graph.

\begin{figure*}[!t]
\centering
\includegraphics[width=.83\textwidth]{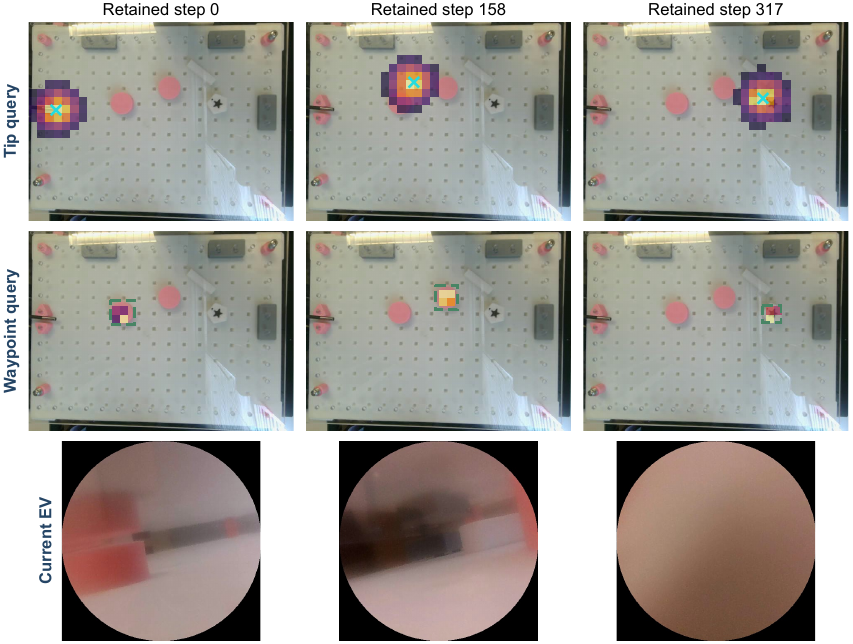}
\caption{Dynamic predictions on a static map in held-out episode L3/0004. Columns show three retained observation indices. Top: tip attention and a cyan cross at the annotated tip. Middle: waypoint attention and its green dashed reference box. Bottom: current EV. Every attention background uses the same initial EBV. Each heatmap is normalized by its own peak for visual comparison of spatial location. Maps are model outputs over original recorded images.}
\label{fig:grounding}
\end{figure*}

\begin{figure}[t]
\centering
\includegraphics[width=\columnwidth]{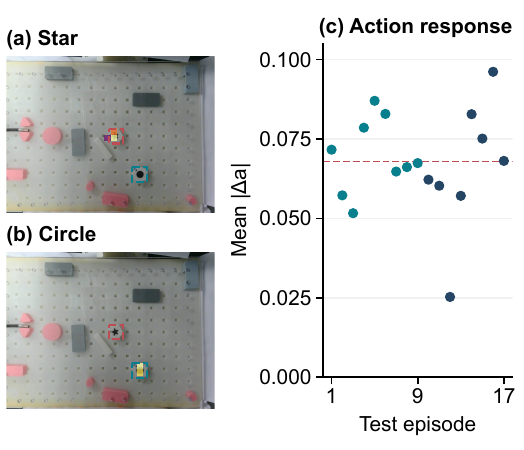}
\caption{Matched-observation instruction intervention. (a,b) Target attention for star and circle instructions on the same L4/0013 test observation; dashed boxes mark the two references, one template-derived. Heatmaps are peak-normalized separately. (c) Mean first-action difference for each dual-target test episode (L4 in teal, L4.5 in navy). The dashed line is the equally weighted episode mean; $\Delta a$ is the normalized first-action difference defined in the text. The measured endpoint is motor-output sensitivity to the instruction.}
\label{fig:language}
\end{figure}

\begin{figure*}[!t]
\centering
\includegraphics[width=.99\textwidth]{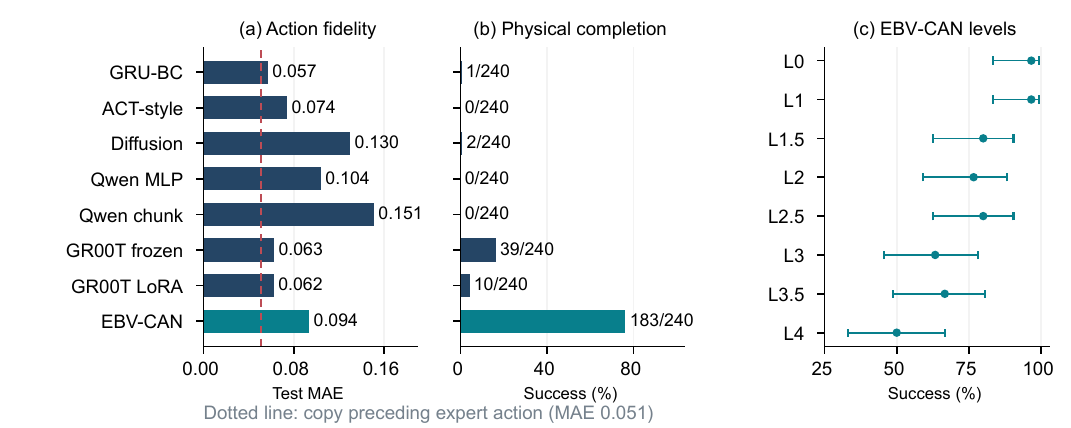}
\caption{Complementary evaluation endpoints for the selected models. (a) First-action MAE over all 7,279 test samples; copying the preceding expert action is a teacher-forced reference. (b) Physical completion for the same model order. (c) EBV-CAN success by task level, with nominal 95\% Wilson intervals over 30 trials per level.}
\label{fig:performance}
\end{figure*}

\begin{figure*}[!t]
\centering\includegraphics[width=.95\textwidth]{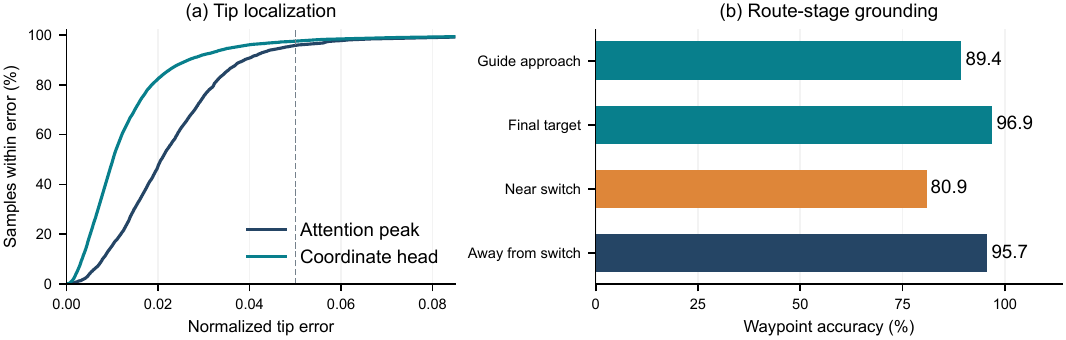}
\caption{Spatial analysis of all 7,279 held-out samples. (a) Empirical cumulative distributions of normalized tip error; the dashed line marks the 0.05 accuracy threshold. (b) Frame-weighted waypoint accuracy for guide versus final-target approach, and for samples near versus away from annotated route changes. Each pair partitions the test set. Near-switch windows comprise five samples before and five starting at each of 46 changes.}
\label{fig:route}
\end{figure*}

\begin{figure*}[!t]
\centering
\includegraphics[width=\textwidth]{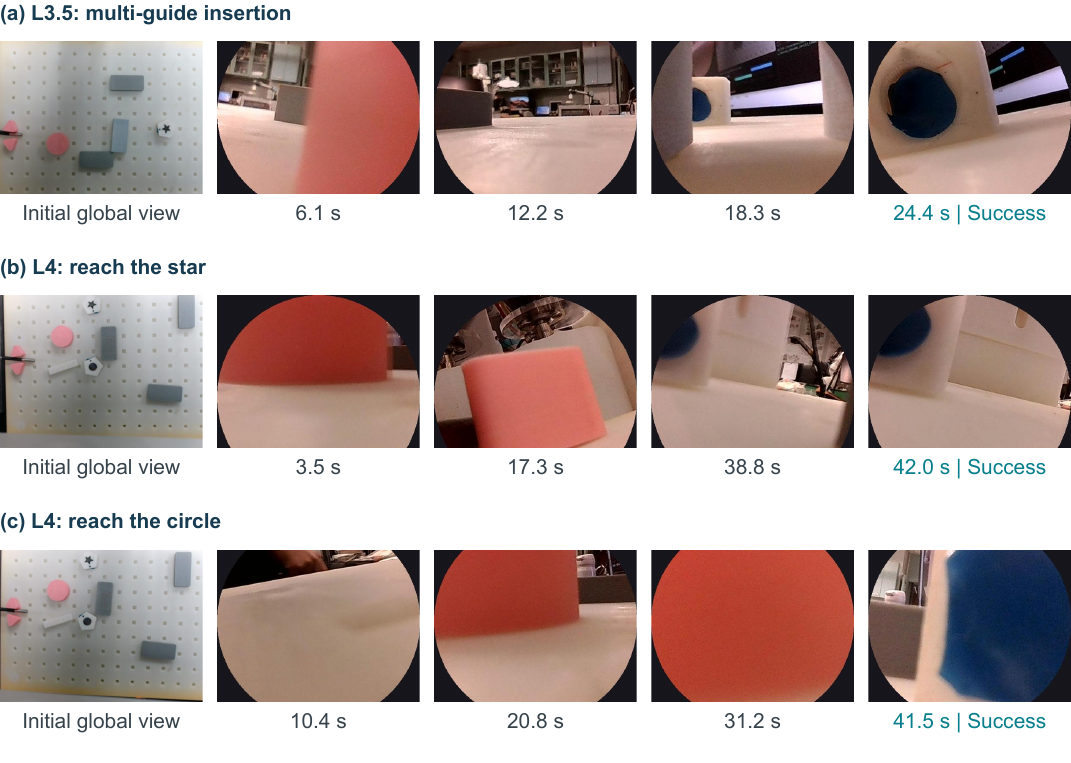}
\caption{Successful EBV-CAN physical executions in demanding task families. (a) L3.5 multi-guide insertion. (b,c) L4 executions instructed to reach the star and circle, respectively. Each row shows the fixed initial global image followed by four original EV observations. Times are relative to the first recorded frame; EV crops emphasize task geometry while retaining the endpoint region. The final images show approach to the blue endpoint region. These selected examples illustrate progression; Table~\ref{tab:physical-baselines} reports the complete outcome counts.}
\label{fig:success}
\end{figure*}

\begin{figure*}[!t]
\centering
\includegraphics[width=\textwidth]{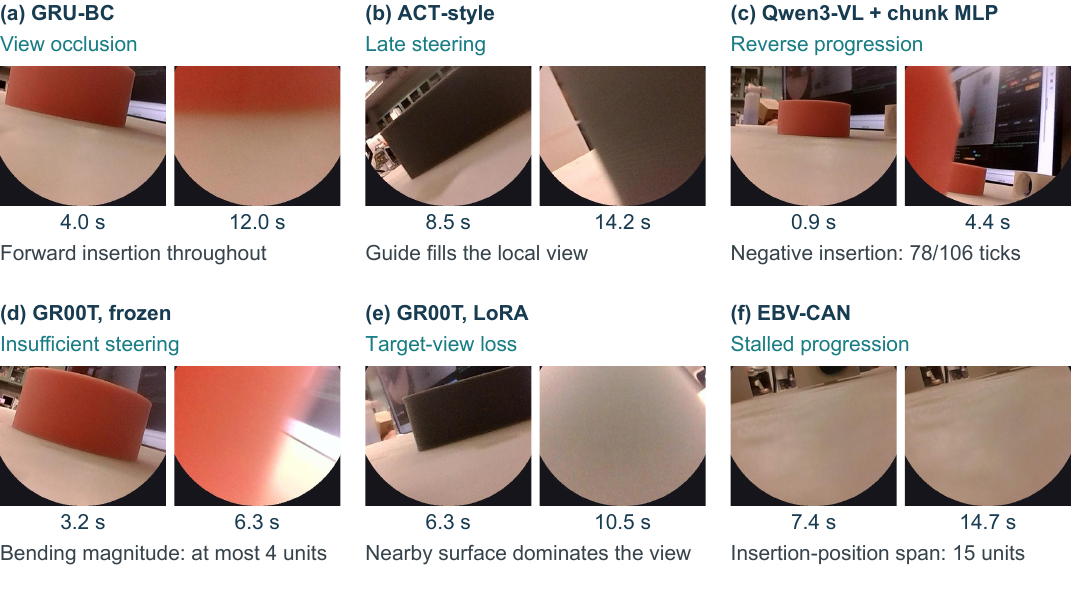}
\caption{Representative L1 failures from original endoscopic recordings, uniformly cropped to emphasize task geometry. Each pair shows an earlier and later observation; time is relative to the first recorded frame. (a) GRU-BC, view occlusion. (b) ACT-style, late steering. (c) Qwen3-VL with a chunk MLP, reverse progression. (d) Frozen GR00T, insufficient steering. (e) GR00T with LoRA, loss of target view. (f) \method{}, stalled progression. Descriptions follow the recorded case notes; images document the visible outcome. SI Sec.~S8 analyzes representative failures from every policy.}
\label{fig:online}
\end{figure*}

The initial $640\times480$ EBV is resized to width 448 and height 336. The frozen DINOv2 small vision transformer with 14-pixel patches (ViT-S/14) produces a $24\times32$ grid of $P=768$ patch features $x_i\in\mathbb{R}^{384}$. Layer normalization (LN), a learned projection, and fixed two-dimensional sinusoidal position encoding form
\begin{equation}
g_i=W_g\operatorname{LN}(x_i)+r_i\in\mathbb{R}^{256}.
\end{equation}
The image and frozen features remain fixed throughout an episode. At inference, map tokens comprise image-derived features and positional encodings.

A width-256 gated recurrent unit (GRU) encodes eight recent EV class tokens, concatenated with the current motor-state embedding. The GRU resets for each finite window, with zero-padded visual history at episode start. A trainable convolutional neural network (CNN) encodes the current EV at $224\times224$. Concatenation and a learned projection fuse these features into $h_t\in\mathbb{R}^{256}$; SI Sec.~S3 specifies the layer dimensions.

\subsection{Language-Conditioned Spatial Queries}
A Qwen-compatible tokenizer \cite{bai2025qwen3}, compact embedding, and text GRU encode the instruction as $z_l$. Vocabulary construction uses the recorded templates and target registry. SI Sec.~S3 specifies the closed-vocabulary encoder and pooling.

For tip, waypoint (wp), and target (tgt), the learned base query $q^0_k$ is conditioned as
\begin{equation}
q^{(0)}_{k,t}=q^0_k+W_h^kh_t+W_l^kz_l.
\end{equation}
Two pre-normalized cross-attention blocks read $g$, each with four-head multi-head attention (MHA) and a feed-forward network (FFN). For block $b$,
\begin{align}
u^{(b)}_t&=q^{(b-1)}_t+\operatorname{MHA}(q^{(b-1)}_t,g,g),\\
q^{(b)}_t&=u^{(b)}_t+\operatorname{FFN}(\operatorname{LN}(u^{(b)}_t)),
\end{align}
where the MHA inputs are layer-normalized internally and the FFN has hidden width 1024. The final block returns head-averaged spatial weights $\alpha_{k,t}\in\mathbb{R}^{768}$. These are supervised prediction maps. The action head uses the \emph{residual query tokens} $q^{(2)}_{k,t}$, which retain query information as well as attention-weighted values.

\subsection{Actions and Training Objectives}
A multilayer perceptron (MLP) maps $[h_t;q^{(2)}_{\mathrm{tip},t};q^{(2)}_{\mathrm{wp},t};q^{(2)}_{\mathrm{tgt},t}]$ through a 512-dimensional hidden layer and $\tanh$ outputs to six two-channel actions. A separate tip-token head predicts normalized coordinates.

The selected training configuration minimizes
\begin{equation}
\begin{split}
\mathcal L={}&\mathcal L_a+0.1\mathcal L_{\rm sm}
+0.25\mathcal L_{\rm tip}+0.25\mathcal L_{\rm wp}\\
&+0.75\mathcal L_{\rm tgt}+\mathcal L_{xy}+0.5\mathcal L_{\rm cf}.
\end{split}
\label{eq:loss}
\end{equation}
The action term is Smooth-L1 loss over valid future commands, masked at episode boundaries. Smoothness is the mean squared difference between adjacent predicted commands within a chunk. Tip attention is trained by cross-entropy to a Gaussian with standard deviation 1.5 patches. Waypoint and target losses are negative log attention mass inside box masks expanded by half a patch for discretization. The coordinate head uses L1 loss.

For dual-target episodes, the counterfactual term swaps star and circle descriptions while holding observations and state fixed, and supervises target-query mass at the other target. Image templates and median target sizes are constructed exclusively from training episodes (SI Sec.~S2.5). Motor-action supervision uses the recorded instruction's expert commands.

Training uses adaptive moment estimation with decoupled weight decay (AdamW) under the shared protocol in SI Sec.~S7: head learning rate $10^{-4}$, effective batch 64, weight decay $10^{-4}$, gradient clipping at 1.0, cosine decay, and a 30-epoch cap. Full-validation first-action MAE selects the checkpoint with patience five. All EBV-CAN analyses use the resulting epoch-11 checkpoint.

\section{Evaluation Protocol}
\begin{table}[t]
\caption{Inference-only interventions on all 7,279 test samples. Accuracy is abbreviated acc.; WP denotes waypoint. Weights remain fixed, isolating inference-time input dependence. Bold indicates highest accuracy or lowest MAE.}
\label{tab:interventions}\centering\small
\setlength{\tabcolsep}{3.5pt}
\begin{tabular}{lrrr}\toprule
Input condition & Tip acc. (\%) & WP acc. (\%) & MAE\\\midrule
Recorded inputs & \textbf{95.78} & \textbf{94.78} & \textbf{0.094}\\
Repeat current EV feature & 92.79 & 94.23 & 0.104\\
Freeze EV at first sample & 22.01 & 43.78 & 0.186\\
Permute global patches & 29.69 & 3.04 & 0.101\\\bottomrule
\end{tabular}
\end{table}

\begin{table*}[!t]
\caption{Held-out first-action MAE for the selected models over 7,279 samples. All models share a 30-epoch cap and validation-MAE selector. Selected epochs start at one; $H$ is the action horizon. Parameter counts are in millions; LoRA deployment uses merged weights. Bold marks the lowest comparable error or parameter count within each block. SI Sec.~S7 gives actual training budgets, raw errors, and curves.}
\label{tab:offline}\centering\small
\setlength{\tabcolsep}{7pt}
\begin{tabular}{lrrrrr}\toprule
Policy & $H$ & Test MAE & Selected epoch & Trainable (M) & Deployed (M)\\\midrule
Zero-action reference & 1 & 0.480 & --- & --- & ---\\
Copy preceding expert action & 1 & \textbf{0.051} & --- & --- & ---\\\midrule
GRU-BC & 1 & \textbf{0.057} & 12 & 2.78 & \textbf{2.8}\\
ACT-style & 16 & 0.074 & 20 & 9.47 & 9.47\\
Diffusion-MLP & 1 & 0.130 & 2 & 3.25 & 3.25\\
Qwen3-VL + MLP & 1 & 0.104 & 14 & 9.84 & 4,447.7\\
Qwen3-VL + chunk MLP & 6 & 0.151 & 25 & \textbf{2.64} & 4,440.5\\
GR00T, frozen & 6 & 0.063 & 24 & 122.89 & 4,560.7\\
GR00T, LoRA & 6 & 0.062 & 16 & 155.92 & 4,560.7\\
\method{} & 6 & 0.094 & 11 & 4.80 & 26.9\\
\bottomrule\end{tabular}

\end{table*}

\begin{table*}[!t]
\caption{Physical success and matched-task offline MAE for the selected checkpoints in SI Sec.~S7. MAE uses 6,271 samples across the eight physical levels; Table~\ref{tab:offline} additionally includes L4.5. Every cell includes 30 trials under the automatic blue-area criterion. Bold marks the lowest MAE and highest success count in each column. GR00T denotes the StarVLA/GR00T implementations.}
\label{tab:physical-baselines}\centering\small
\setlength{\tabcolsep}{3pt}
\begin{tabular}{@{}lr*{8}{c}r@{}}
\toprule
& & \multicolumn{9}{c}{Physical success (automatic blue-area criterion)}\\
\cmidrule(lr){3-11}
Policy & Offline MAE & L0 & L1 & L1.5 & L2 & L2.5 & L3 & L3.5 & L4 & All\\
\midrule
GRU-BC & \textbf{0.056} & 1/30 & 0/30 & 0/30 & 0/30 & 0/30 & 0/30 & 0/30 & 0/30 & 1/240 (0.42\%)\\
ACT-style & 0.073 & 0/30 & 0/30 & 0/30 & 0/30 & 0/30 & 0/30 & 0/30 & 0/30 & 0/240 (0.00\%)\\
Diffusion-MLP & 0.129 & 1/30 & 0/30 & 0/30 & 0/30 & 0/30 & 1/30 & 0/30 & 0/30 & 2/240 (0.83\%)\\
Qwen3-VL + MLP & 0.102 & 0/30 & 0/30 & 0/30 & 0/30 & 0/30 & 0/30 & 0/30 & 0/30 & 0/240 (0.00\%)\\
Qwen3-VL + chunk MLP & 0.148 & 0/30 & 0/30 & 0/30 & 0/30 & 0/30 & 0/30 & 0/30 & 0/30 & 0/240 (0.00\%)\\
GR00T, frozen & 0.061 & 12/30 & 6/30 & 7/30 & 6/30 & 2/30 & 1/30 & 3/30 & 2/30 & 39/240 (16.25\%)\\
GR00T, LoRA & 0.061 & 0/30 & 3/30 & 3/30 & 0/30 & 0/30 & 0/30 & 0/30 & 4/30 & 10/240 (4.17\%)\\
\midrule
\method{} & 0.092 & \textbf{29/30} & \textbf{29/30} & \textbf{24/30} & \textbf{23/30} & \textbf{24/30} & \textbf{19/30} & \textbf{20/30} & \textbf{15/30} & \textbf{183/240 (76.25\%)}\\
\bottomrule
\end{tabular}

\end{table*}
\subsection{Offline Prediction and Spatial Grounding}
Action prediction uses all 7,279 samples in the 50 test episodes. MAE is measured for the first predicted action, both averaged over normalized channels and separately in controller units. The preceding action is the demonstrated preceding action in this test. A copy-previous-action reference quantifies temporal persistence under this teacher-forced protocol.

Comparisons include GRU-BC, an ACT-style transformer, a task-specific Diffusion-MLP, frozen Qwen3-VL with single-action or chunk MLP heads, and StarVLA/GR00T with frozen or LoRA-adapted backbones \cite{hu2021lora}. All models use the same episode partitions, effective batch 64, 30-epoch cap, and full-validation first-action MAE selector, with patience five. AdamW uses head learning rate $10^{-4}$ and LoRA rate $10^{-5}$. SI Sec.~S7 gives the architectures, actual optimization budgets, per-epoch curves, and final results. Offline and physical comparisons use these selected models.

Tip localization is correct when the Euclidean distance between the attention-peak patch center and the normalized annotated tip is less than 0.05. Waypoint localization is correct when its peak center falls within the unexpanded waypoint box. Target and guide mass use the expanded training masks. Top-area intersection-over-union (IoU) selects as many highest-weight patches as there are patches in the reference mask. Guide-only statistics use 2,082 samples; target statistics use all 7,279. The metrics quantify localization in normalized image coordinates and agreement with annotated object regions.

\subsection{Matched-Observation Language Test}
The 17 held-out dual-target episodes contain 2,145 samples. Each is evaluated twice: with the recorded instruction and with star/circle descriptions swapped. Images, history, positions, and preceding action remain identical. Strict query switching requires target attention to favor the recorded target under the original instruction and the alternate target under the swapped instruction. We also measure
\begin{equation}
\Delta a_t=\tfrac12\|\hat A_{t,1}(l)-\hat A_{t,1}(l')\|_1.
\end{equation}
We summarize $\Delta a_t$ per episode and bootstrap the 17 episode means with 10,000 resamples. Episode-level resampling accounts for within-episode frame correlation. This paired inference test measures output dependence on language; its endpoint is the change in the predicted command.

\subsection{Physical Trials}

The physical evaluation comprises 1,920 trials of the eight selected checkpoints from SI Sec.~S7: 240 per policy and 30 per policy at L0, L1, L1.5, L2, L2.5, L3, L3.5, and L4. The completed evaluation register supplies the level-wise counts. L4.5 contributes to offline evaluation. Repeated trials share task configurations, so sessions and layouts define the clustering structure.

All eight policies use the automatic blue-area endpoint criterion: a trial succeeds when the largest blue connected component reaches at least 5,000 pixels in any one of the final three EV frames. This endpoint measures visible approach to the blue target region. Table~\ref{tab:physical-baselines} retains 30 trials in every policy--level cell. Nominal 95\% Wilson intervals summarize the observed proportions. Representative L1 failures supply qualitative examples and behavioral descriptions (SI Sec.~S8).

\section{Results}
\subsection{Dynamic Localization on a Fixed Image}
The selected checkpoint achieves 95.8\% tip-peak accuracy, 97.5\% tip-coordinate accuracy, and 94.8\% waypoint-peak accuracy. Mean normalized tip-peak error is 0.024. Target attention mass is 1.000 and guide-only mass is 0.893; corresponding top-area IoU values are 0.580 and 0.519. These complementary measures characterize probability concentration inside the labeled region and spatial overlap with its extent.

Figure~\ref{fig:grounding} shows successive predictions from a held-out three-guide demonstration. The background remains fixed while predicted tip and waypoint locations change with EV evidence and motor state. Mean waypoint Jensen--Shannon divergence over pairs ten retained observations apart is 0.054 (6,779 pairs), quantifying temporal change. Waypoint accuracy separately evaluates spatial agreement. Later recorded overhead images in SI Fig.~S4 illustrate guide-supported shaft shapes and provide annotation context.

\subsection{Localization Across Route Stages}
Route-stage evaluation resolves spatial performance within the held-out trajectories (Fig.~\ref{fig:route}). The coordinate head achieves mean normalized tip error 0.014, median 0.010, and 95th percentile 0.037. The equally weighted episode mean is 0.014, with a 95\% episode-bootstrap interval [0.012, 0.015]. Its continuous output refines the patch-grid estimate, whose mean error is 0.024.

Waypoint accuracy is 89.4\% over 2,082 guide-directed samples and 96.9\% over 5,197 final-target samples. The held-out sequences contain 46 annotated waypoint changes in 33 episodes. Across the five retained samples before and five starting at each change, accuracy is 80.9\% (460 samples), compared with 95.7\% over the remaining 6,819 samples. These transition windows comprise 6.3\% of test samples but account for 88 of 380 waypoint errors (23.2\%). The remaining-route annotations reveal a concentrated difficulty: selecting the next spatial objective as the route advances. This supports evaluating transition windows explicitly when developing contact-aided controllers. Equal episode weighting gives overall waypoint accuracy 95.3\% [92.3, 97.6].

\subsection{Sensitivity to Visual Inputs}
We intervene on the inputs while holding the selected checkpoint fixed (Table~\ref{tab:interventions}). Repeating the current EV feature at all valid history positions changes action MAE from 0.094 to 0.104 and tip-peak accuracy from 95.78\% to 92.79\%. Freezing both EV branches at the first observation reduces tip and waypoint accuracy to 22.01\% and 43.78\%, while motor state remains available. These results assign complementary roles to the observations: the cached EBV defines a spatial reference, and evolving EV evidence updates the estimated configuration within it. Recent visual history provides additional temporal information beyond the current feature.

Permuting cached global patch features while keeping position encodings fixed reduces waypoint accuracy to 3.04\% and increases action MAE to 0.101. The permutation preserves the set of visual features while disrupting their spatial assignment. This dissociation distinguishes supervised grounding from action sensitivity: the navigation skip and residual queries provide additional information paths to the action head. The result supports separate evaluation of spatial predictions, action error, and closed-loop behavior when assessing the global representation.

\subsection{Language Changes Attention and Actions}
Strict query switching occurs in all 2,145 dual-target samples under the derived-box protocol. Figure~\ref{fig:language} gives a matched-observation example and first-action differences for all 17 episodes. The frame-weighted mean normalized difference is 0.067; the equally weighted episode mean is 0.068, with bootstrap 95\% interval [0.060, 0.075]. Mean absolute changes in insertion and bending commands are 1.236 and 10.871 controller units.

Changing only the instruction changes the first motor command. The paired action response complements strict target-query switching, which measures compliance with the trained derived-box protocol. Physical target-switch effectiveness requires paired rollouts with alternate-target outcome judgments.

\subsection{Action Fidelity and Model Size}
Table~\ref{tab:offline} and Fig.~\ref{fig:performance} report the selected models' first-action errors. GRU-BC obtains the lowest learned-policy MAE, 0.057; \method{} obtains 0.094, and Diffusion-MLP obtains 0.130. Copying the preceding expert action gives 0.051, quantifying temporal persistence under teacher forcing.

\method{} uses 4.80 million trainable and approximately 26.9 million deployed parameters, including its frozen visual encoder. Task learning concentrates in the navigation and semantic-query modules, with feedback incorporated after each executed command.

\subsection{Physical Performance and Offline Rankings}
Table~\ref{tab:physical-baselines} pairs physical success with the corresponding selected checkpoint's offline MAE on the 6,271 test samples from the eight physical task levels. \method{} achieves 183/240 successes (76.25\%), frozen GR00T achieves 39/240 (16.25\%), and LoRA achieves 10/240 (4.17\%). GRU-BC and Diffusion-MLP achieve 1/240 and 2/240; ACT-style and both Qwen heads each achieve 0/240. \method{} has the highest observed success at every level and exceeds the strongest baseline by 60.0 percentage points overall.

The lowest imitation error and highest physical completion belong to different policies. GRU-BC obtains MAE 0.056 with 0.42\% physical success; \method{} obtains 0.092 with 76.25\% success. The GR00T variants also obtain lower MAE than \method{}, while achieving lower physical completion. Thus accurate first-action prediction at expert states and reliable execution along policy-induced trajectories are empirically distinct properties. The common training cap and validation selector align model selection; the architectures, pretrained representations, and spatial objectives remain explicit differences.

\subsection{Successful Progression Through Difficult Tasks}
Figure~\ref{fig:success} shows successful L3.5 and L4 executions. In the insertion-dependent L3.5 case, the EV progresses from a nearby guide surface through an open view to the target, while the policy retains the same initial global image. The L4 cases reach the blue endpoint region under star and circle instructions in scenes containing both target identities. These sequences show completion despite large changes in local appearance, complementing the held-out evidence that evolving EV inputs update global spatial estimates. They also illustrate the physical setting in which target-conditioned predictions operate. The separate L4 runs describe successful execution under their recorded instructions; the matched-observation language test isolates instruction dependence quantitatively.

\subsection{Zero-Success Cells and Representative Failures}
The completed matrix contains 42 zero-success cells out of 64: seven for GRU-BC, eight each for ACT-style and the two Qwen heads, six for Diffusion-MLP, and five for LoRA GR00T. Each 0/30 denotes 30 recorded attempts below the endpoint criterion. Its nominal Wilson 95\% interval is 0--11.4\%; an aggregate 0/240 gives 0--1.6\%. SI Sec.~S5 lists every zero-valued level and the corresponding behavioral evidence. These intervals describe finite trial counts under repeated task setups.

Figure~\ref{fig:online} and SI Sec.~S8 connect the outcomes to concrete behaviors. ACT-style examples pass the target or steer before or after the required bend. GRU-BC examples combine view occlusion, incorrect steering, and target pass-by; the Diffusion-MLP cases show incorrect or premature steering. Qwen's single-action head exhibits excessive and incorrectly directed bending, while its chunk head also produces reverse progression. Frozen GR00T cases include insufficient or delayed steering; LoRA cases include incorrect steering and loss of target view. These observations illustrate how motor commands and evolving observations jointly shape an unsuccessful trajectory: steering direction determines guide approach, timing determines the support configuration, and visibility determines the next feedback available.

\method{} retains failures involving stalled progression, incorrect or delayed bending, and target-view loss. Its success rate decreases from 29/30 at L0/L1 to 15/30 at L4, locating the remaining difficulty in more demanding navigation and target-selection tasks. The L1 examples connect failure outcomes to visible steering and progression errors, motivating recovery behaviors that restore informative views and useful support configurations. Supplementary Videos 1--7 document acquisition and physical executions (SI Sec.~S6).

\raggedbottom
\section{Discussion}
The results establish spatially supervised feedback control with fixed global context. EV interventions and held-out grounding connect evolving local observations to changing global spatial estimates, while 183 physical completions establish navigation performance. Matched-observation instruction swaps change both target attention and motor output, linking semantic intent to the control computation.

Action MAE and physical completion evaluate different state distributions. Teacher-forced prediction visits demonstrated states and supplies the expert's preceding action, explaining the 0.051 copy-previous reference. Physical execution generates new observations and contact configurations through policy commands \cite{ross2011reduction}. GRU-BC's lowest MAE and EBV-CAN's highest completion rate demonstrate the consequence: action fidelity alone gives an incomplete ranking of navigation performance.

VLCA makes route progress measurable through canonical IDs, ordered plans, and 495 remaining-route updates. Transition windows contain 6.3\% of samples and 23.2\% of waypoint errors, identifying a focused target for improved progress estimation. Object boxes locate support geometry; remaining-route labels identify which object matters next. Their linkage to actions supports learning the timing of sequential support transitions.

The evidence also defines the scope of the present system. Occlusion, stalled progression, target pass-by, and steering errors persist in physical execution. GRU-BC's lowest imitation error and 1/240 physical successes expose the evaluation mismatch between expert-state prediction and policy-induced trajectories. Current sensing consists of images and motor signals; force and shaft-shape measurements are important extensions for resolving contact mechanics and supporting safety-aware clinical translation. Validation covers planar, rigid guide structures, stable scene registration, two control DoFs, and shared layout families. Deformable three-dimensional anatomy therefore forms a distinct validation stage, with additional mechanical and sensing demands.

Future work will connect these failure modes to targeted recovery training. Recovery demonstrations and corrective imitation will address view occlusion, target pass-by, and stalled progression, while route-transition objectives will emphasize the exact timing of steering onset and release. A visibility-aware protocol will trigger automatic retreat and view reacquisition when the target view is lost. Force and shaft-shape measurements will inform recovery-state estimation, and deformable three-dimensional routing environments will extend validation beyond the present platform. A recursive self-improvement (RSI) development loop, informed by Endoscopic Recursive Self-Improvement (EndoRSI)~\cite{ng2026endorsi}, will use physical failure evidence to revise demonstrations, objectives, and policies, followed by held-out physical evaluation of each revision.

\FloatBarrier
\section{Conclusion}
We presented \method{} and VLCA for language-conditioned contact-aided endoscope navigation. An initial global image supplies persistent context, while local observations and motor state update spatial predictions and actions. VLCA links 517 demonstrations and 74,693 action samples to dense spatial and remaining-route supervision. Shared-protocol training and 1,920 physical trials establish distinct offline and physical rankings: \method{} achieves 76.25\% success, while the lowest-MAE policy achieves 0.42\%. Representative failures connect unsuccessful execution to steering, visibility, and recovery challenges. The combined evidence supports evaluating action fidelity, spatial progress, and physical completion together.

\section*{Acknowledgment}
The authors thank Huxin Gao for introducing the concept of contact-aided navigation for robotic endoscopes, and Zhiqing Tang for helping design and print the EBV camera support.

ChatGPT (OpenAI) assisted with language revision and icon generation. The authors reviewed the resulting material and remain responsible for its scientific content.

\raggedbottom
\putbib[references]

\end{bibunit}
\clearpage

\setcounter{section}{0}
\setcounter{subsection}{0}
\setcounter{figure}{0}
\setcounter{table}{0}
\setcounter{equation}{0}
\renewcommand{\thesection}{S\arabic{section}}
\renewcommand{\thesubsection}{S\arabic{section}.\arabic{subsection}}
\renewcommand{\thesubsectiondis}{\thesubsection.}
\renewcommand{\thefigure}{S\arabic{figure}}
\renewcommand{\thetable}{S\arabic{table}}
\renewcommand{\theequation}{S\arabic{equation}}
\twocolumn[{
\centering
{\LARGE Supplementary Information\par}
\vspace{0.5em}
{\Large EBV-CAN: Vision--Language Grounding for Contact-Aided Endoscope Navigation\par}
\vspace{1.5em}
}]
\begin{bibunit}[IEEEtran]
\raggedbottom
This supplement specifies the Vision-Language Contact-Aided (VLCA) dataset's acquisition and annotation structure, model configuration, selected checkpoints, metric definitions, physical-trial records, and multimedia. All model results use the shared training protocol and validation-selected checkpoints in Sec.~S7.

\section{Platform and Task Families}
The platform uses an industrial flexible endoscope with a short actively bending distal section and a long passive shaft. A motorized bending knob controls one planar bending degree of freedom; counter-rotating friction rollers insert or retract the shaft. We use DJI RoboMaster M2006 P36 motors. Commands use controller scales 20 for insertion and 150 for bending. All command magnitudes are expressed in controller units.

An overhead camera records the external bird's-eye view (EBV) of the planar board, and a tip-mounted illuminated camera records the endoscopic view (EV). Demonstrations are teleoperated with an Xbox controller. The measured channels are camera images and motor signals. Contact-related interpretation concerns visible shaft support, motion, and image-space route progress.

\begin{figure*}[t]
\centering\includegraphics[width=.89\textwidth]{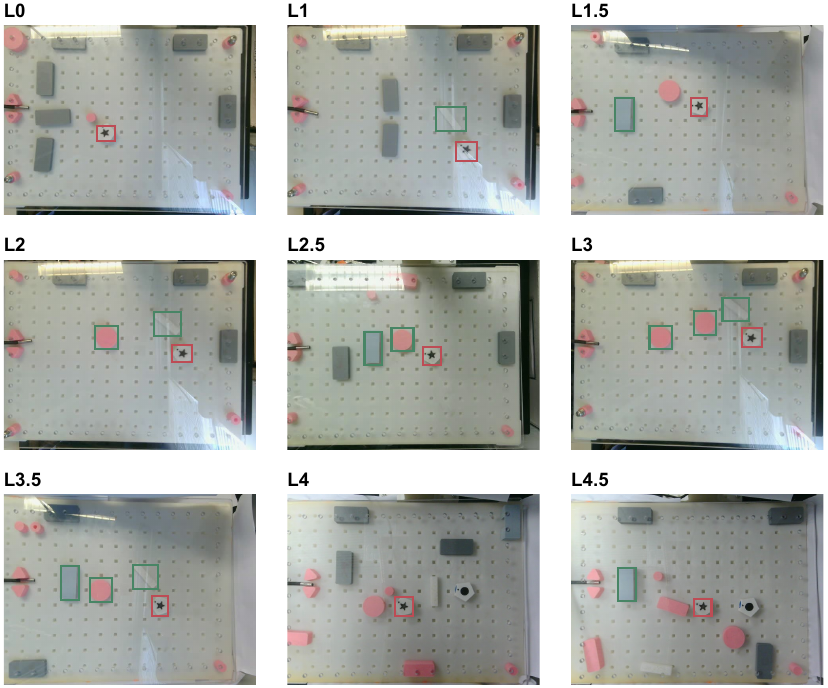}
\caption{Original initial EBV examples for all offline task families. Green boxes identify guides appearing in each episode's plan; red identifies the instructed target. Each level identifies a family of route, insertion, and target-selection requirements.}
\label{fig:levels}
\end{figure*}

Figure~\ref{fig:levels} illustrates the nine task families. Planned guide counts are zero for L0, one for L1/L1.5, two for L2/L2.5, and three for L3/L3.5. L4 contains both target identities and zero to two planned guides; among its 93 policy episodes, 73 follow direct-to-target plans, 15 use one guide, and five use two. All 90 L4.5 plans use one guide. The .5 suffix denotes task designs emphasizing insertion-dependent traversal. Both motor channels are available across the task families, and the plan identifies the selected guides among the visible blocks.

The policy observation contract is one initial global static EBV per episode. Both cameras record $640\times480$ images. The EV is postprocessed using a centered $480\times480$ square crop, and an inscribed circular mask. Training and physical inference use this same view transformation, followed by the encoder-specific resizing in Sec.~S7. Later overhead observations support demonstration annotation. Online metadata marks the policy EBV as frozen.

\section{VLCA Data and Annotations}
\subsection{Corpus, Filtering, and Partitions}
There are 518 deposited recordings. Filtering for a nonempty action export excludes L4/0066, leaving 517 episodes and 74,693 action samples. All retained samples have nonzero action and both tip and next-waypoint supervision. Spatial annotation contains 94,173 frame records in the broader recording set, while the policy corpus contains the 74,693 retained action samples. The deposited image archive contains 92,488 EV and 92,495 EBV files. The annotation export retains 1,678 additional historical frame keys beyond the deposited EBV image set. All 75,191 distinct image files referenced by the policy table are present.

\begin{figure*}[t]
\centering\includegraphics[width=.95\textwidth]{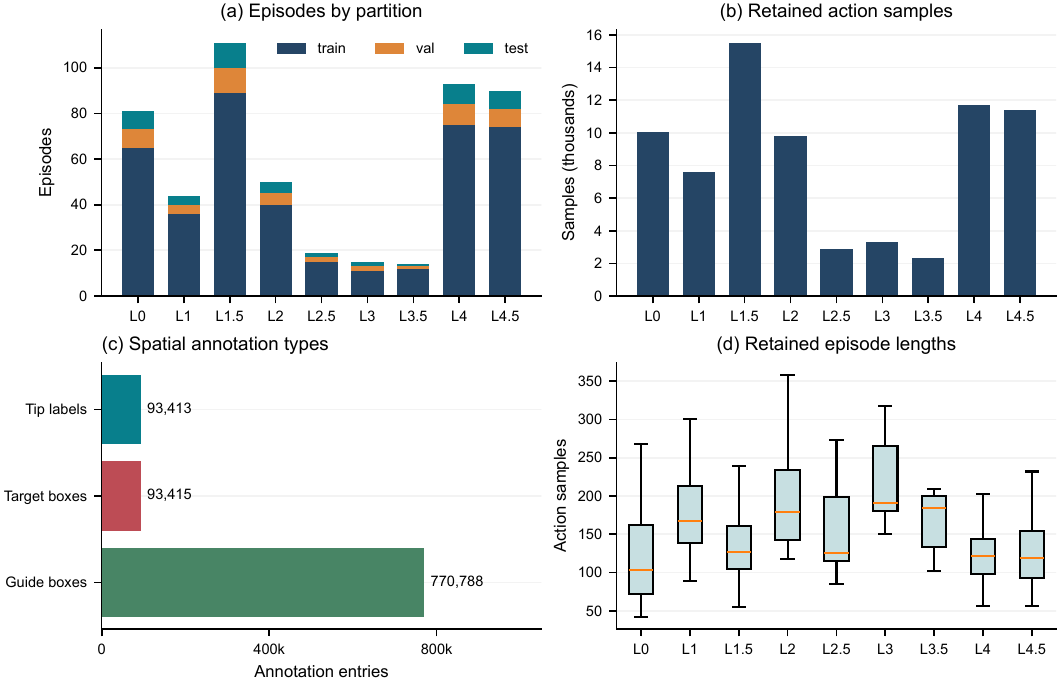}
\caption{VLCA composition. (a) Episode partitions by task family. (b) Retained action samples. (c) Spatial annotation entries across the raw annotation export; tip annotations are counted once, with the coordinate field treated as an alternative representation. (d) Retained episode lengths, showing medians, quartiles, and whiskers based on 1.5 times the interquartile range (IQR), with outliers omitted from the plot.}
\label{fig:stats}
\end{figure*}

Table~\ref{tab:split} gives the episode-disjoint, level/target-stratified partition (seed 42). Layouts are shared across the recorded distribution; the partition holds out whole episodes. The 183 dual-target policy episodes comprise 93 L4 and 90 L4.5 episodes. L4 contains 48 circle and 45 star episodes; L4.5 contains 45 of each. The held-out dual-target subset has 17 episodes and 2,145 samples. Figure~\ref{fig:stats} summarizes dataset scale and imbalance.

\begin{table*}[t]
\caption{Exact policy partition sizes, reported as episodes / action samples.}
\label{tab:split}\centering\small
\begin{tabular}{lrrrr}\toprule
Level & Training & Validation & Test & Total\\\midrule
L0 & 65 / 8,451 & 8 / 749 & 8 / 880 & 81 / 10,080\\
L1 & 36 / 6,312 & 4 / 588 & 4 / 713 & 44 / 7,613\\
L1.5 & 89 / 12,815 & 11 / 1,284 & 11 / 1,439 & 111 / 15,538\\
L2 & 40 / 7,687 & 5 / 1,066 & 5 / 1,084 & 50 / 9,837\\
L2.5 & 15 / 2,268 & 2 / 231 & 2 / 390 & 19 / 2,889\\
L3 & 11 / 2,361 & 2 / 425 & 2 / 509 & 15 / 3,295\\
L3.5 & 12 / 2,054 & 1 / 176 & 1 / 119 & 14 / 2,349\\
L4 & 75 / 9,460 & 9 / 1,118 & 9 / 1,137 & 93 / 11,715\\
L4.5 & 74 / 9,392 & 8 / 977 & 8 / 1,008 & 90 / 11,377\\\midrule
All & 417 / 60,800 & 50 / 6,614 & 50 / 7,279 & 517 / 74,693\\\bottomrule
\end{tabular}
\end{table*}

Median adjacent retained-sample spacing is 0.050~s; 2.622\% of intervals exceed 0.075~s, with maximum 2.205~s. Complete eight-observation windows have median duration 0.351~s, 95th percentile 0.604~s, and maximum 2.956~s. Window size is defined in retained samples. Evaluation preserves the observed sequence and its acquisition gaps.

\subsection{Episode and Sample Schema}
Each episode links its two image streams, action export, instruction, target identity, and spatial annotations in JavaScript Object Notation (JSON). The processed table preserves episode identifier (ID), split, task level, retained frame index, timestamp, image paths, motor positions, normalized current action, preceding action, tip coordinates, and next-waypoint box/ID. Episode metadata stores the initial EBV and instructed-target box.

The annotation JSON contains the following linked fields:
\begin{itemize}
\item \texttt{labels}: per-frame entries with class, normalized geometry, and proposal-provenance flags.
\item \texttt{tipPoints}: normalized tip positions, derived from a point marker or the center of a robot-tip box.
\item \texttt{referenceBlocks}: canonical guide boxes with identifiers such as B1, B2, and B3.
\item \texttt{globalPlanningList}: the ordered episode plan ending in \texttt{Target}.
\item \texttt{framePlanningLists}: per-frame remaining plans used to select the next waypoint.
\item \texttt{guidingConfig}: the distance-based guide-progress rule and its threshold.
\end{itemize}
The normalized box origin is its upper-left corner. Image coordinates increase rightward and downward. The point field and robot-tip entry describe the same underlying tip annotation.

\subsection{Annotation Workflow}
\begin{figure*}[t]
\centering\includegraphics[width=.98\textwidth]{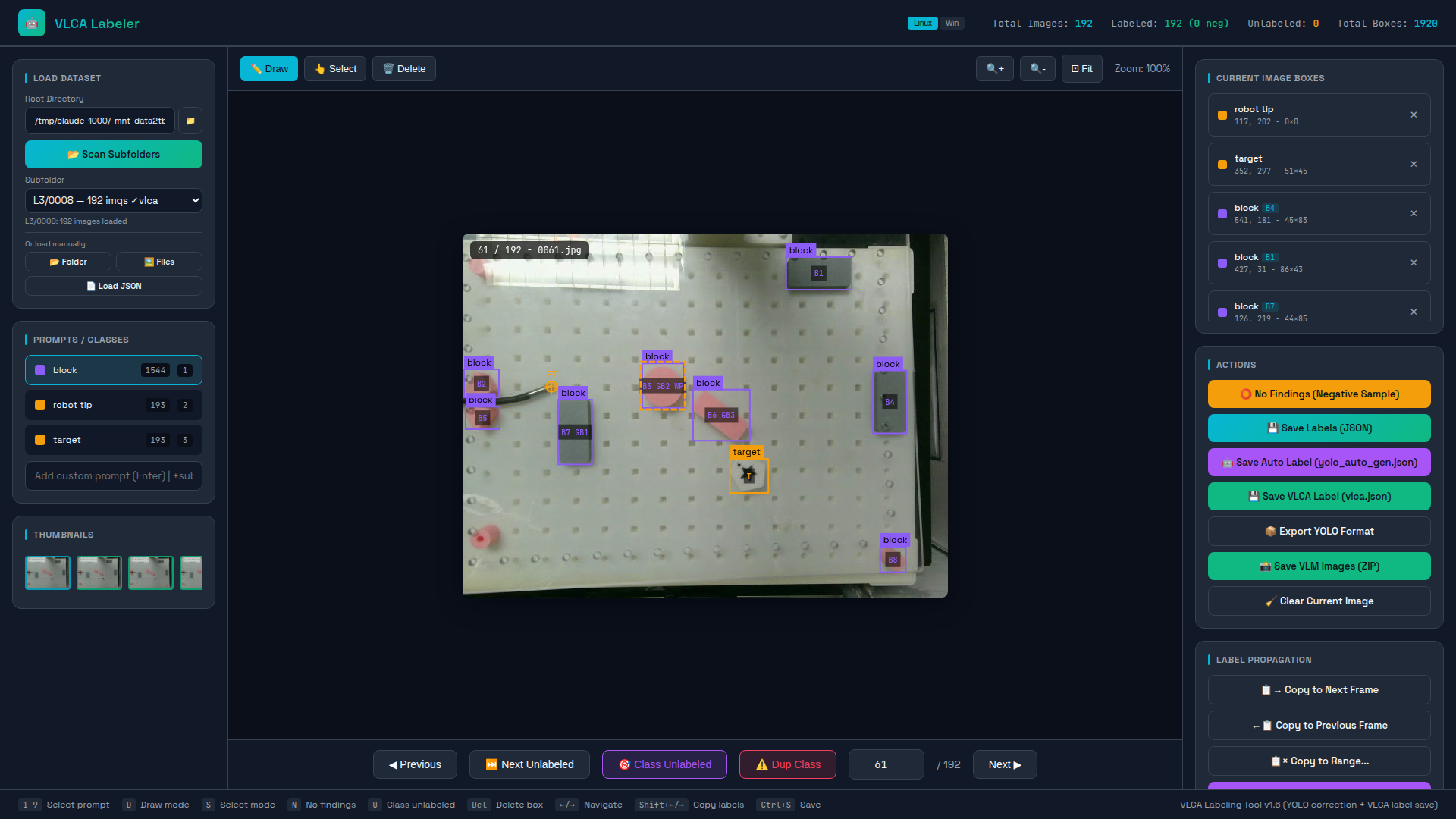}
\caption{The proposed VLCA YOLO-backboned annotation platform and the user interface. The central view links tip, object, target, and waypoint annotations; side panels expose editing, propagation, and JSON export. The interface illustrates the annotation workflow and stored field relationships; VLM denotes vision-language model.}
\label{fig:labelui}
\end{figure*}

The annotation tool supports You Only Look Once (YOLO) detector proposals, manual box and point editing, canonical block assignment, route ordering, and frame-to-frame label propagation. Figure~\ref{fig:labelui} shows the original interface. 

A direct census of all deposited annotation files finds 94,173 labeled-frame records, 93,413 tip-point records, 93,085 frame-plan records, and 957,616 spatial entries. The entries comprise 770,788 block boxes, 93,415 target boxes, and 93,413 robot-tip entries. Each denominator refers to its named field. The current policy subset has complete tip/waypoint coverage after filtering. The export additionally contains 4,041 canonical reference boxes, 518 ordered plans, and 495 remaining-plan updates. Tip coordinates and frame plans coexist on 93,083 frames. Among remaining-plan records, 25,709 identify a guide as the next waypoint and 67,376 identify the target.

For progress labeling, the tool computes distance from the annotated tip to a guide's box edge in pixels and compares it with a threshold normalized by the larger image dimension. A reached guide is removed from the remaining list for subsequent frames; the final target remains in the list. Thresholds are $1/32$ in 437 recordings, 0.03 in 79, and $3/64$ in two. At $640\times480$, these correspond to 20, 19.2, and 30 pixels. Proximity to a box defines the image-space route-progress label.

\subsection{Instructions, Causal State, and Release}
The processed corpus has two vision-language-action (VLA) instruction strings. Both contain a target name and description followed by a shared generic navigation directive. Episode-specific routes are stored in the ordered-plan fields. The exact strings are reproduced below.
\noindent\textbf{star:} [VLA]
You are a continuum robot navigation expert. Your task is to navigate the endoscope toward the target star, described as the five-point star-shaped target. Use the guiding blocks when contact with them is needed to change the navigation direction or expand the reachable workspace. Ground the endoscopic tip and the next waypoint from the visual observations, and generate the next robot action.

\noindent\textbf{circle:} [VLA]
You are a continuum robot navigation expert. Your task is to navigate the endoscope toward the target circle, described as the circular disk-shaped target. Use the guiding blocks when contact with them is needed to change the navigation direction or expand the reachable workspace. Ground the endoscopic tip and the next waypoint from the visual observations, and generate the next robot action.

State is
\begin{equation}
s_t=\left[
\operatorname{clip}(p^{\rm ins}_t/20),
\operatorname{clip}(p^{\rm bend}_t/150),
a^{\rm ins}_{t-1},a^{\rm bend}_{t-1}\right],
\end{equation}
with clipping to $[-1,1]$ and zero preceding action at episode start. These quantities encode clipped motor state; global tip position is a separate image-space annotation. In the policy export the bending channel is named legacy M3; the standardized motor protocol maps planar bending to physical M4. The manuscript names physical functions to avoid confusing these conventions.

For each training sample, the current state is repeated across the eight visual-feature positions. The state input is the current four-dimensional vector at every history position. Future actions are output labels only. The action chunk has a validity mask at episode ends; the smoothing regularizer still applies across all predicted chunk positions. These partitions retain the recorded two-template instruction setting and episode-disjoint evaluation protocol.

The compact review deposit uses LeRobot v3.0 with separate train/validation/test datasets, retaining 417/50/50 episodes and 60,800/6,614/7,279 action samples. Typed sample tables preserve motor values, instruction and annotation fields, and source-image identities. A mapping connects split-local episode indices to the original episode keys. The annotation directory contains all 518 final label files and paired full-recording videos. Videos use H.264 compression with constant rate factor (CRF) 18. The regular 20-Hz storage clock accompanies the original acquisition timestamps. Policy EV images are $480\times480$. Reported model metrics use the original image collection; the compact release preserves numerical supervision and source correspondence.

\subsection{Derived Alternate-Target Boxes}
Each episode's target annotation identifies the instructed object. For L4/L4.5, the alternate-target box is generated from class-averaged templates using normalized cross-correlation on the initial EBV, suppressing the instructed-target neighborhood. Template dimensions are median annotated widths and heights from the 417 training episodes. Crop collection accepts up to 60 valid training crops per class.

The fixed templates supply alternate boxes for counterfactual training and held-out evaluation. Strict query switching measures compliance with this derived-reference protocol. Validation and test episodes contribute only to evaluation.

\subsection{Additional Structured-Language Annotation Exports}
Two complementary exports reuse paired EBV/EV observations. The \emph{plan} export supplies an overlay-enhanced EBV and raw EV, with structured answers for normalized tip coordinates, an ordered global plan, and the next waypoint ID. The \emph{wild} export supplies raw EBV/EV images, with answers for tip coordinates and a next-waypoint bounding box. Both retain target descriptions, frame and episode IDs, the remaining plan, a guide-proximity distance and threshold, a critical-frame flag, and a list of acceptable next waypoints. This connects scene recognition, route ordering, and local spatial grounding in a machine-readable form.

The structured-language partition comprises 415 training and 103 evaluation episodes, covering the 518 deposited recordings (Table~\ref{tab:vlmarchive}). The supervised fine-tuning (SFT) subset contains 300 selected frames from 300 of the training episodes. The reinforcement fine-tuning (RFT) training and evaluation subsets contain 15,770 and 3,915 critical frames, respectively, according to the guide-proximity flag. Counts apply separately to the two supervision formats over shared observations. Table~\ref{tab:split} reports the separate 417/50/50 action-policy partition over 517 episodes.

\begin{table}[t]
\caption{Structured-language annotation exports for 518 recordings. Counts apply separately to both plan and wild formats. The SFT subset draws from the 415 training episodes.}
\label{tab:vlmarchive}\centering\small
\begin{tabular}{lrrr}\toprule
Subset & Episodes & Frames & Critical frames\\\midrule
SFT training & 300 & 300 & 0\\
RFT training & 415 & 73,169 & 15,770\\
Evaluation & 103 & 19,015 & 3,915\\\bottomrule
\end{tabular}
\end{table}

\section{Model and Optimization Details}
Table~\ref{tab:model} records the selected configuration of external bird's-eye-view contact-aided navigation (EBV-CAN). Its frozen self-distillation with no labels, version 2 (DINOv2), encoder uses a small vision transformer with 14-pixel patches (ViT-S/14) to provide width-384 visual features \cite{oquab2023dinov2}. The current-EV convolutional neural network (CNN) has successive channel widths 32, 64, 128, 256, and 256, using strided convolutions, group normalization, rectified linear units (ReLU), adaptive pooling, and a width-256 output. Inputs use red, green, and blue (RGB) channels in $[0,1]$. Navigation and text each use a gated recurrent unit (GRU); cross-attention includes a feed-forward network (FFN). The action head is a multilayer perceptron (MLP) using layer normalization, a 512-unit Gaussian error linear unit (GELU) layer, dropout 0.05, and a 12-output $\tanh$ layer. The tip head uses 128 hidden units and a sigmoid two-coordinate output. Optimization uses adaptive moment estimation with decoupled weight decay (AdamW).

\begin{table}[t]
\caption{\method{} architecture and training settings. Spatial weights refer to tip, waypoint (wp), and target (tgt). The shared training protocol is detailed in Sec.~S7.}
\label{tab:model}\centering\small
\begin{tabular}{ll}\toprule
Component & Setting\\\midrule
EBV input / patch grid & $448\times336$ / $24\times32$\\
EV feature input & $336\times336$\\
EV window / action horizon & 8 / 6 retained samples\\
Current-EV CNN input & $224\times224$\\
Motor state / embedding & 4 / 64 dimensions\\
Navigation GRU & width 256; reset each window\\
Text embedding / GRU & 128 / 256 dimensions\\
Cross-attention & 2 residual blocks; 4 heads\\
Query width / FFN width & 256 / 1024\\
AdamW learning rate & $10^{-4}$\\
Weight decay / batch size & $10^{-4}$ / 64\\
Training seed & 0\\
Maximum epochs / patience & 30 / 5\\
Gradient norm clip & 1.0\\
Spatial weights (tip, wp, tgt) & 0.25, 0.25, 0.75\\
Coordinate / counterfactual & 1.0 / 0.5\\
Within-chunk smoothness & 0.1\\\bottomrule
\end{tabular}
\end{table}

The compact text vocabulary is constructed from all recorded instruction strings and the target registry. A tokenizer compatible with Qwen3 vision-language (Qwen3-VL) supplies IDs; the embedding and GRU are trained for this navigation task. Final query tokens contain the residual, cross-attention, and feed-forward updates. These complete tokens, together with the navigation feature, form the implemented action-head input.

Training minimizes masked Smooth-L1 action loss, squared within-chunk differences, Gaussian tip-attention cross-entropy, box-mass losses, L1 coordinate loss, and counterfactual target mass. For normalized box $(x,y,w,h)$, the mass mask includes patch centers in
\begin{equation}
[x-\tfrac1{64},x+w+\tfrac1{64}]
\times[y-\tfrac1{48},y+h+\tfrac1{48}].
\end{equation}
The tip Gaussian has standard deviation 1.5 patches. Counterfactual loss runs a second forward pass only for dual-target samples with a valid derived box. The counterfactual target is spatial attention mass; action targets are the recorded expert commands.

\section{Offline Evaluation}
\subsection{Selected Model}
Full-validation first-action mean absolute error (MAE) selects the EBV-CAN checkpoint at epoch 11 (epochs start at one).
All spatial grounding, instruction swaps, route-stage analysis, and input interventions use this checkpoint and the training-only alternate-target references. The same selected model supplies the physical comparison. Test first-action MAE is 0.094, with insertion and bending errors 1.945 and 13.547 controller units.

\subsection{Metric Definitions and Per-Level Results}
For peak index $j$, the normalized patch center is
$((j\bmod32+0.5)/32,(\lfloor j/32\rfloor+0.5)/24)$.
Tip accuracy uses Euclidean error $<0.05$. The coordinate-head accuracy uses the same threshold. Waypoint accuracy requires the peak center to be inside the \emph{unexpanded} reference box. Box mass and intersection-over-union (IoU) use the half-patch-expanded masks above.

For mask $M$ and attention $\alpha$, top-area IoU selects the $|M|$ highest-weight patches as set $Q$ and computes $|Q\cap M|/|Q\cup M|$. Table~\ref{tab:offlinelevel} gives per-level results for the selected model. Values are frame-weighted means.

\begin{table*}[t]
\caption{Selected-model results by held-out task family. Accuracy (acc.) is in percent; normalized (norm.) MAE averages both channels; raw action errors use controller units.}
\label{tab:offlinelevel}\centering\small
\begin{tabular}{lrrrrrr}\toprule
Level & Samples & Tip-peak acc. & Waypoint acc. & Norm. MAE & Insertion MAE & Bending MAE\\\midrule
L0 & 880 & 93.18 & 100.00 & 0.077 & 1.605 & 11.158\\
L1 & 713 & 88.08 & 99.30 & 0.081 & 1.788 & 10.985\\
L1.5 & 1,439 & 98.33 & 98.75 & 0.084 & 1.596 & 13.370\\
L2 & 1,084 & 97.32 & 86.44 & 0.075 & 1.443 & 11.739\\
L2.5 & 390 & 100.00 & 96.92 & 0.090 & 1.815 & 13.391\\
L3 & 509 & 97.45 & 88.21 & 0.078 & 1.648 & 10.989\\
L3.5 & 119 & 99.16 & 95.80 & 0.064 & 1.399 & 8.702\\
L4 & 1,137 & 95.16 & 90.85 & 0.148 & 3.180 & 20.522\\
L4.5 & 1,008 & 96.03 & 97.12 & 0.102 & 2.262 & 13.700\\
\midrule All & 7,279 & 95.78 & 94.78 & 0.094 & 1.945 & 13.547\\
\bottomrule\end{tabular}

\end{table*}

Waypoint temporal change uses natural-log Jensen--Shannon divergence (JSD) between predictions ten retained samples apart within the same episode, with Kullback--Leibler (KL) divergence $D_{\rm KL}$:
\begin{equation}
\begin{split}
\operatorname{JSD}(p,q)&=\tfrac12D_{\rm KL}(p\Vert m)+\tfrac12D_{\rm KL}(q\Vert m),\\
m&=(p+q)/2.
\end{split}
\end{equation}
It is 0.054 across 6,779 pairs. The corresponding physical interval follows the recorded sample timestamps.

For original and alternate boxes $B_o,B_d$, strict switching requires
\begin{equation}
M_o(B_o)>M_o(B_d),\qquad M_s(B_d)>M_s(B_o),
\end{equation}
where $M_o,M_s$ are target-query masses under original and swapped instructions. All 2,145 dual-target samples satisfy this derived-reference criterion. First-action sensitivity holds all other model inputs fixed, including the preceding expert action. The frame mean normalized L1 difference is 0.067 and median 0.054. Equal episode weighting gives 0.068; a percentile bootstrap over 17 episode means (10,000 resamples) gives [0.060, 0.075]. This interval describes the predicted first-action response to the instruction swap.

\subsection{Inference-Only Input Interventions}
All interventions use the same epoch-11 weights and 7,279 test samples. Repeating the current EV feature replaces each valid history token by the current token, preserving the padding mask, current-image CNN input, and motor state. Freezing EV uses the first retained EV feature and image from each episode for both local branches, while motor state and history-validity masks continue to change. Global permutation applies one fixed random permutation to the 768 frozen patch features before their learned projection and fixed positional encoding. Local observations, state, and language remain unchanged.

The main paper reports all three interventions, including the global permutation: tip/waypoint accuracies fall to 29.69\%/3.04\%, while normalized first-action MAE is 0.101 versus 0.094 with recorded inputs. Spatial assignment is perturbed while the set of patch contents is preserved. This result demonstrates a separation between supervised spatial readouts and teacher-forced action error. All values describe inference-time input dependence under fixed weights.

\subsection{Baseline Interpretation}
Gated recurrent unit behavior cloning (GRU-BC) is a recurrent cloning implementation. An action chunking transformer (ACT)-style policy predicts 16 actions, and a diffusion policy with an MLP denoiser (Diffusion-MLP) predicts one action. These are task-specific implementations motivated by their respective method families \cite{mandlekar2021matters,zhao2023learning,chi2025diffusion}. Four-billion-parameter (4B) Qwen3-VL features are frozen for the MLP baselines. The StarVLA implementations of Generalist Robot 00 Technology (GR00T) use a flow-matching action expert and horizon six; low-rank adaptation (LoRA) additionally adapts the backbone \cite{bai2025qwen3,community2026starvla,bjorck2025gr00t,hu2021lora}. Section~S7 gives all training recipes and selected results.

Reported deployment counts include resident frozen encoders. EBV-CAN has 4,804,686 trainable and 26,861,262 deployed parameters. Execution applies the first command of each chunk before new feedback.

\section{Physical Evaluation and Failure Evidence}
\subsection{Observations, Execution, and Checkpoints}
\begin{figure*}[t]
\centering\includegraphics[width=.98\textwidth]{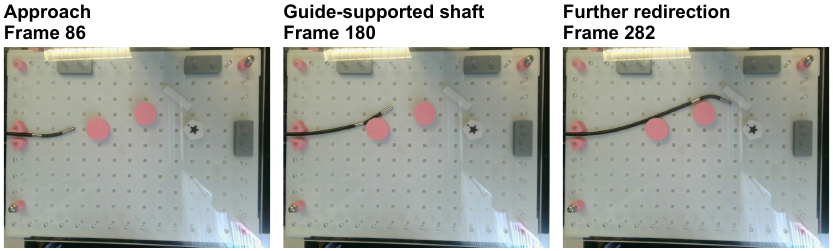}
\caption{Recorded overhead frames from demonstration L3/0004 showing approach and guide-supported shaft redirection. These original images provide qualitative mechanical context. The sequence comes from the demonstration annotation stream; attention plots use the fixed initial image.}
\label{fig:contact}
\end{figure*}

Figure~\ref{fig:contact} shows evolving overhead images from teleoperated demonstration L3/0004. These original images document guide-supported shaft configurations. Deployment instead receives one frozen initial EBV and ongoing EV observations.

The completed evaluation register reports 1,920 trials, with 240 for each selected checkpoint and 30 for each of eight task levels. All policies use the selected checkpoints in Sec.~S7. Representative L1 failures support qualitative interpretation of the level-wise outcomes.

\subsection{Endpoint and Trial Proportions}
All eight policies use the automatic blue-area endpoint criterion: a trial succeeds when the largest blue connected component reaches at least 5,000 pixels in any one of the final three EV frames. This endpoint measures visible approach to the blue target region. Each success/trial pair in main-paper Table~V comes from the completed level-wise register. The numerator counts endpoint-positive trials, and the denominator includes all recorded attempts. Recorded case descriptions provide a separate source of behavioral interpretation.

Table~\ref{tab:online} reports \method{} outcomes by task level. Wilson intervals use $z=1.96$ and the exact numerator and denominator. They are nominal binomial summaries: repeated sessions and layouts introduce dependence between attempts, so the intervals describe the retained counts.

\begin{table}[t]
\caption{Automatic blue-area physical outcomes for \method{} and nominal Wilson 95\% intervals. Bold marks the highest observed rate across levels.}
\label{tab:online}\centering\small
\begin{tabular}{lrrr}\toprule
Level & Positive / total & Rate (\%) & Interval (\%)\\\midrule
L0 & \textbf{29/30} & \textbf{96.7} & 83.3--99.4\\
L1 & \textbf{29/30} & \textbf{96.7} & 83.3--99.4\\
L1.5 & 24/30 & 80.0 & 62.7--90.5\\
L2 & 23/30 & 76.7 & 59.1--88.2\\
L2.5 & 24/30 & 80.0 & 62.7--90.5\\
L3 & 19/30 & 63.3 & 45.5--77.9\\
L3.5 & 20/30 & 66.7 & 48.8--80.8\\
L4 & 15/30 & 50.0 & 33.2--66.8\\\midrule
All & 183/240 & 76.25 & 70.5--81.2\\\bottomrule
\end{tabular}
\end{table}

\subsection{Model-Specific Failure Modes}
The failure sequences distinguish steering, visibility, and progression errors. GRU-BC advances toward a guide until the surface occludes the EV; its other examples show incorrect steering and target pass-by. ACT-style reaches the vicinity of the target or guide but applies steering too early, too late, or in the wrong direction. Diffusion-MLP similarly exhibits incorrect or premature steering. These trajectories illustrate the sensitivity of guide approach to steering direction and timing along policy-induced states.

The Qwen3-VL single-action head exhibits large bending commands followed by loss of the target view. Its chunk head additionally produces sustained reverse progression. Frozen GR00T shows weak or delayed steering while insertion continues, bringing a guide surface into the local view. LoRA-adapted GR00T exhibits incorrect or mistimed steering and target-view loss. Adaptation and action chunking therefore coexist with distinct physical failure mechanisms in these evaluated implementations.

EBV-CAN's failures include stalled progression, incorrect or late steering, and target-view loss. Repeatedly revisiting similar EV appearances while the actuator state changes indicates a recovery problem: useful progress requires a coordinated change in insertion and bending that restores an informative view. These observations motivate steering-transition supervision, corrective demonstrations, and explicit retreat-and-reacquisition behavior. The sequences provide behavioral evidence; contact forces and shaft shape would further resolve the underlying mechanics.

\subsection{Interpretation of Every Zero-Success Cell}
Table~\ref{tab:zero-cells} covers all 42 zero-valued cells in main-paper Table~V. Each cell is 0/30, with nominal Wilson interval 0--11.35\%. ACT-style and both Qwen heads also have 0/240 aggregate outcomes, with interval 0--1.58\%. Frozen GR00T and \method{} have positive counts at every evaluated level. GRU-BC succeeds once at L0, Diffusion-MLP succeeds once each at L0 and L3, and LoRA succeeds at L1, L1.5, and L4.

\begin{table*}[t]
\caption{All zero-success cells and relevant L1 case evidence. The listed levels identify observed 0/30 outcomes; behavioral examples originate from L1.}
\label{tab:zero-cells}\centering\small
\setlength{\tabcolsep}{4pt}
\begin{tabular}{@{}lp{.34\textwidth}p{.39\textwidth}@{}}\toprule
Policy & Levels with 0/30 & Representative recorded behaviors\\\midrule
GRU-BC & L1, L1.5, L2, L2.5, L3, L3.5, L4 & Occlusion, incorrect steering, pass-by, target-view loss\\
ACT-style & L0, L1, L1.5, L2, L2.5, L3, L3.5, L4 & Pass-by, incorrect steering, early/late steering\\
Diffusion-MLP & L1, L1.5, L2, L2.5, L3.5, L4 & Incorrect steering, early steering\\
Qwen3-VL + MLP & L0, L1, L1.5, L2, L2.5, L3, L3.5, L4 & Target-view loss, early/incorrect steering\\
Qwen3-VL + chunk MLP & L0, L1, L1.5, L2, L2.5, L3, L3.5, L4 & Reverse progression, target-view loss\\
GR00T, LoRA & L0, L2, L2.5, L3, L3.5 & Incorrect/early/insufficient steering, target-view loss\\\bottomrule
\end{tabular}
\end{table*}

The zero-valued L0 cells concern direct-to-target navigation, showing that successful target approach already differentiates these deployments before planned guide use. The L1 cases expose steering and visibility failures during the first guide interaction. Zero-valued higher-guide cells additionally require successful progression through a sequence of support configurations. At L4, endpoint completion combines geometric traversal with the instructed target task. These distinctions motivate targeted diagnostics for target acquisition, support transitions, and recovery. The L1 observations establish concrete behavioral hypotheses for the other levels; their frequency and causal attribution remain level-specific quantities.

\subsection{Matched-Checkpoint Offline Comparisons}
Table~\ref{tab:context} reports first-action MAE for the same selected checkpoints on every physical task level. All policies use identical offline rows. The normalized metric averages insertion error divided by 20 and bending error divided by 150. The 6,271-row aggregate weights each level by its offline sample count. Physical proportions weight 30 attempts at every level. Main-paper Table~IV additionally includes 1,008 L4.5 rows. This explicit weighting separates task coverage from checkpoint identity.

\begin{table*}[t]
\caption{Selected-checkpoint offline MAE by physical task level. GR00T F/L denote frozen/LoRA variants. Bold marks the lowest MAE per row. Physical outcomes for the same weights appear in main-paper Table~V.}
\label{tab:context}\centering\small
\setlength{\tabcolsep}{4pt}
\begin{tabular}{@{}lr*{8}{r}@{}}\toprule
Level & $n$ & GRU & ACT & Diffusion & Qwen MLP & Qwen chunk & GR00T F & GR00T L & \method{}\\\midrule
L0 & 880 & \textbf{0.046} & 0.061 & 0.123 & 0.090 & 0.142 & 0.048 & 0.052 & 0.077\\
L1 & 713 & \textbf{0.052} & 0.062 & 0.124 & 0.082 & 0.132 & 0.053 & 0.055 & 0.081\\
L1.5 & 1439 & \textbf{0.049} & 0.066 & 0.122 & 0.104 & 0.140 & 0.056 & 0.054 & 0.084\\
L2 & 1084 & \textbf{0.044} & 0.056 & 0.121 & 0.085 & 0.137 & 0.048 & 0.050 & 0.075\\
L2.5 & 390 & \textbf{0.060} & 0.081 & 0.127 & 0.106 & 0.148 & 0.062 & 0.061 & 0.090\\
L3 & 509 & \textbf{0.046} & 0.069 & 0.118 & 0.082 & 0.129 & 0.051 & 0.048 & 0.078\\
L3.5 & 119 & \textbf{0.047} & \textbf{0.047} & 0.119 & 0.099 & 0.130 & 0.048 & 0.049 & 0.064\\
L4 & 1137 & \textbf{0.091} & 0.113 & 0.160 & 0.145 & 0.194 & 0.099 & 0.096 & 0.148\\
\bottomrule\end{tabular}

\end{table*}

GRU-BC gives the lowest eight-level aggregate MAE (0.056) and 1/240 physical successes. The corresponding values are 0.061 and 39/240 for frozen GR00T, 0.061 and 10/240 for LoRA, and 0.092 and 183/240 for \method{}. The comparison uses the selected-model predictions throughout. All reported comparisons use the selected models in Sec.~S7.

\section{Multimedia}
Supplementary Videos 1--7 retain the following index:
\begin{enumerate}
\item L0 directly reachable teleoperated demonstration.
\item L3 multi-guide teleoperated demonstration.
\item L4.5 target-conditioned, insertion-dependent demonstration.
\item Successful L4 star-target closed-loop trial.
\item Successful L3.5 multi-guide insertion trial.
\item Successful L4 circle-target closed-loop trial.
\item Unsuccessful L3 closed-loop trial with incorrect steering.
\end{enumerate}
Videos 1--3 show sampled overhead views from demonstration and annotation at 12 frames/s. Videos 4--7 pair the fixed initial global image with the original EV stream, resampled at 20~Hz using recorded timestamps. Their visible elapsed-time labels follow the source frames. The successful physical clips correspond to main-paper Fig.~9. Physical outcomes use the automatic blue-area endpoint criterion.

\section{Training Protocol and Selected Models}
\subsection{Shared Training Protocol}
The comparison gives every learned policy the same 417 training episodes, 50 validation episodes, and 50 test episodes. An epoch visits all 60,800 training rows once in shuffled order. Effective batch 64 gives 950 optimizer updates per epoch. Every model receives a maximum of 30 epochs, or 28,500 updates and 1,824,000 row exposures, with validation after each complete epoch. Actual budgets follow the same stopping rule: five consecutive validation checks with improvement below $10^{-4}$. The stored best checkpoint minimizes the full-validation first-action normalized MAE across all 6,614 validation rows. Reported means weight individual rows, including the final validation minibatch.

AdamW uses $(0.9,0.999)$, weight decay $10^{-4}$, gradient-norm clipping at 1.0, and cosine decay over the 30-epoch cap. Task heads use learning rate $10^{-4}$; Qwen LoRA adapters use $10^{-5}$. Seed zero initializes task modules before optimization and sets training order. Compact policies use float32 tensors with TensorFloat-32 (TF32) matrix operations enabled; the large-backbone policies use bfloat16 autocasting. GRU-BC accumulates four microbatches of 16; LoRA accumulates eight microbatches of eight; the other implementations use batch 64 directly. LoRA weights each microbatch by its valid-position count within the complete 64-row batch, then clips gradients and updates once. A masked-loss oracle verifies this full-batch normalization. Frozen Qwen features and DINOv2 features preserve their pretrained representations. These feature caches encode individual observations independently. Trainable visual CNNs use bilinearly resized $224\times224$ RGB inputs scaled to $[0,1]$; the StarVLA image pair also enters its processor at $224\times224$. The frozen DINOv2 representation follows the EBV patch and EV class-token processing in Sec.~S3. Input preprocessing uses fixed transforms throughout training and evaluation.

All policies receive the recorded current EV, initial EBV, and causal motor state appropriate to their architecture. Numeric state branches scale positions and preceding commands by the 20/150 controller scales; the frozen Qwen-feature prompt retains the four recorded causal-state fields as JSON. The language-conditioned policies additionally receive the recorded instruction. Histories end at the current observation, and supervised action chunks stay within episode boundaries. The image preflight verifies 75,191 distinct source files and resolves relocated paths before optimization. Frozen encoders, action horizons, language interfaces, and the additional spatial supervision of \method{} remain explicit architectural differences. The comparison controls data partitions, the optimization cap, and the checkpoint-selection rule while evaluating these complete task implementations.

Validation and test inference use separate fixed random seeds for the stochastic policies. The test partition is evaluated after validation selects the checkpoint. Training logs record the epoch, optimizer updates, optimization-row exposures, training objective, validation MAE, learning rate, and elapsed time. This experiment reports one training seed; the curves characterize optimization for that seed. All runs use an NVIDIA GeForce RTX~5090 with Compute Unified Device Architecture (CUDA) 12.8. Compact runs use PyTorch 2.10.0; StarVLA runs use a PyTorch 2.12.0 development build, Transformers 4.57.0, and Parameter-Efficient Fine-Tuning (PEFT) 0.19.1.

\begin{table*}[t]
\caption{Training with a common optimization cap and validation selector. Epochs start at one. Validation uses 6,614 rows and test uses 7,279 rows; raw MAEs use controller units. Loss scales are model-specific. Bold marks the lowest validation or test error in its column.}
\label{tab:controlled-training}\centering\small
\setlength{\tabcolsep}{4pt}
\begin{tabular}{lrrrrrrr}\toprule
Policy & \shortstack{Epochs\\trained / selected} & Updates & \shortstack{Final\\train loss} & \shortstack{Validation\\MAE} & \shortstack{Test\\MAE} & \shortstack{Insertion\\MAE} & \shortstack{Bending\\MAE}\\\midrule
GRU-BC & 17 / 12 & 16,150 & 0.007 & \textbf{0.056} & \textbf{0.057} & \textbf{1.133} & 8.640\\
ACT-style & 25 / 20 & 23,750 & 0.045 & 0.069 & 0.074 & 1.553 & 10.579\\
Diffusion-MLP & 7 / 2 & 6,650 & 0.125 & 0.128 & 0.130 & 2.491 & 20.354\\
Qwen3-VL + MLP & 19 / 14 & 18,050 & 0.018 & 0.100 & 0.104 & 2.555 & 12.082\\
Qwen3-VL + chunk MLP & 30 / 25 & 28,500 & 0.047 & 0.148 & 0.151 & 3.236 & 21.077\\
GR00T, frozen & 29 / 24 & 27,550 & 0.025 & 0.058 & 0.063 & 1.355 & 8.623\\
GR00T, LoRA & 17 / 16 & 16,150 & 0.033 & 0.061 & 0.062 & 1.389 & \textbf{8.302}\\
\method{} & 16 / 11 & 15,200 & 0.932 & 0.093 & 0.094 & 1.945 & 13.548\\
\bottomrule\end{tabular}

\end{table*}

\subsection{Model-Specific Training Recipes}
The GRU, ACT-style, and diffusion policies use separate EV and EBV CNN weights with the same five-stage encoder structure. Channel widths are 32, 64, 128, 256, and 256; stride is two throughout. The first convolution uses a $5\times5$ kernel and the others use $3\times3$ kernels. Each stage applies eight-group normalization and ReLU, followed by global average pooling and a 256-dimensional linear projection. Smooth-L1 objectives use transition parameter one.

\paragraph{GRU-BC}
Two trainable CNNs encode EV and EBV into 256-dimensional features. A state MLP produces 64 features, and a width-256 GRU feeds a 128-unit action head with a two-dimensional $\tanh$ output. Training minimizes Smooth-L1 first-action loss over eight-observation causal windows. The GRU resets at each window, and a validity mask skips history padding. Training and evaluation use this same finite-history computation.

\paragraph{ACT-style chunk transformer}
The task implementation encodes EV, EBV, and state into three width-256 tokens. Four transformer encoder layers and four decoder layers use four attention heads, 1,024-unit feed-forward blocks, and dropout 0.1. Sixteen learned queries predict a 16-action chunk. The objective is mean squared action error over valid future positions. First-action MAE on the held-out validation episodes selects the checkpoint. The ACT-style row uses this deterministic task implementation~\cite{zhao2023learning}.

\paragraph{Diffusion-MLP}
Two visual CNNs and a state MLP produce a width-256 condition. A denoiser with three 512-unit hidden layers predicts noise for a two-dimensional action, conditioned on the observation and a width-256 sinusoidal time embedding followed by a two-layer sigmoid linear unit (SiLU) projection. Training uses 100 cosine-schedule diffusion steps and mean squared noise-prediction error. Inference uses 20 correctly spaced deterministic denoising diffusion implicit model (DDIM) updates \cite{song2020denoising}. Each update uses the cumulative noise levels at both the current and next selected time. The clean-action estimate is clipped to $[-1,1]$, and the noise direction is recomputed consistently after clipping. A numerical oracle check recovers prescribed clean actions exactly under the implemented schedule. The diffusion row evaluates the stated sampler together with the shared training and selection protocol.

\paragraph{Qwen3-VL MLP heads}
The frozen 4B backbone supplies a 2,560-dimensional final-token feature from the original-resolution EBV/EV pair, instruction, and causal-state prompt \cite{bai2025qwen3}. The saved pretrained image processor supplies its adaptive image resizing and normalization. The feature is taken after the final root mean square (RMS) normalization, at the last actual token under right padding. Recomputing 64 evenly spaced cache rows gives a minimum cosine similarity of 0.999 and maximum relative L2 difference of 0.036 under bfloat16 batch computation. This explicit feature definition resolves differences between intermediate hidden-state hooks across library versions. The single-action head uses layer normalization, hidden widths 2,560 and 1,280, GELU, dropout 0.05, and a two-output $\tanh$ layer. The chunk head uses a 1,024-unit hidden layer and twelve outputs reshaped into six actions. Both optimize Smooth-L1 action error; the chunk version adds squared adjacent-action differences with weight 0.1. A validity mask excludes action-supervision positions beyond each episode's end. The adjacent-command regularizer covers the predicted six-action chunk.

\paragraph{StarVLA/GR00T and LoRA}
The Qwen3-VL-4B backbone conditions a 12-layer base-size diffusion transformer (DiT-B) flow-matching action expert with a six-action horizon \cite{community2026starvla,bjorck2025gr00t}. The task expert initializes from the same seed-zero weights in the frozen and LoRA runs. Post-normalization multimodal token features condition the expert. The state branch maps four inputs through 1,024 hidden units to the action-transformer width of 768. Flow matching draws four noise/time samples per observation and minimizes squared velocity error. Its retained time schedule draws $u\sim\mathrm{Beta}(1.5,1)$ and sets $t=(0.999-\min(u,0.999))/0.999$. Future-position masks respect episode boundaries. Inference uses four Euler updates and clips commands to the normalized controller limits. The frozen variant trains the action expert using cached post-normalization tokens stored as exact bfloat16 bit patterns; a cache round-trip check gives zero reconstruction error. The train/validation cache is constructed before optimization, and test features are encoded after checkpoint selection. The LoRA run recomputes backbone features with gradients at every update. Four central processing unit (CPU) workers prepare image tensors and tokens; a 64-row check gives exact equality with the retained input processor for token IDs, attention masks, image pixels, and image grids. The LoRA variant additionally trains rank-16 adapters with scale 32 and dropout 0.05 on the language model's query, key, value, output, gate, up, and down projections \cite{hu2021lora}. Gradient checks verify that the loss reaches the adapter parameters. Full-partition validation selects the checkpoint. Deployment merges the LoRA weights, giving 4,560.7 million resident parameters; the 33.03 million adapter parameters contribute to training.

\paragraph{EBV-CAN}
The architecture, DINOv2 features, six-action horizon, and spatial-loss weights follow Table~\ref{tab:model}; the shared optimizer and action-MAE selector govern this run. Alternate-target templates and their median crop sizes are constructed exclusively from the 417 training episodes, using up to 60 labeled crops per target class. These fixed templates generate alternate-target references for subsequent partitions. The recorded instruction templates are identical in the training and full corpora.

\subsection{Optimization Results and Interpretation}

\begin{figure*}[p]
\centering\includegraphics[width=\textwidth]{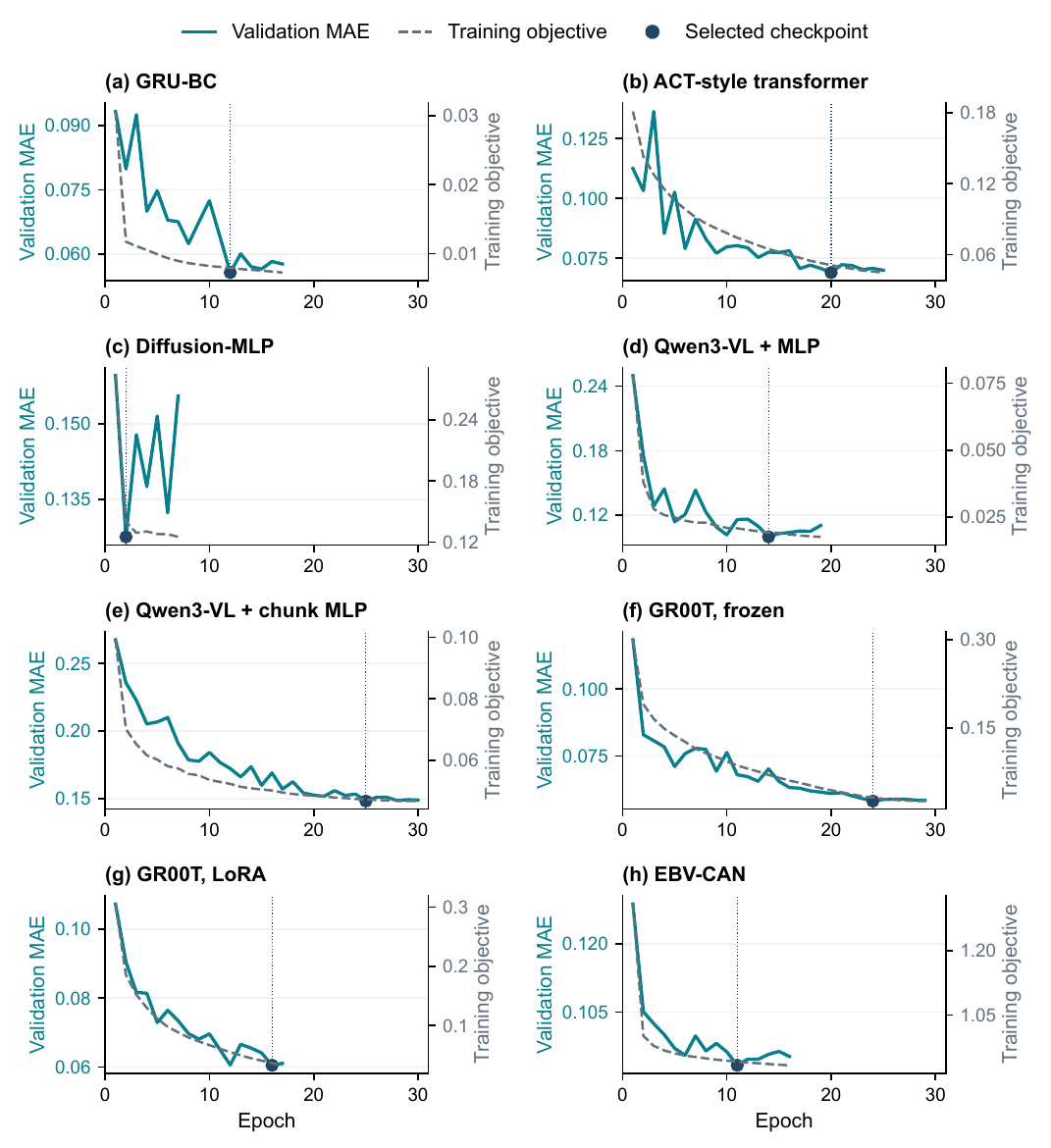}
\caption{Optimization trajectories under the shared protocol. Each panel shows the epoch-mean training objective and complete-validation first-action MAE on separately labeled axes. The marked validation point identifies the stored best checkpoint. Each epoch contains 950 optimizer updates and 60,800 optimization-row exposures. Objective scales follow each model's learning formulation.}
\label{fig:training-curves}
\end{figure*}

Table~\ref{tab:controlled-training} records 7--30 completed epochs across the eight runs. Figure~\ref{fig:training-curves} identifies the selected validation point on each trajectory, connecting test performance to optimization history. GRU-BC and ACT-style achieve test MAEs 0.057 and 0.074; EBV-CAN achieves 0.094 and Diffusion-MLP achieves 0.130.

The frozen and LoRA GR00T runs complete 29 and 17 epochs, with selected test MAEs 0.063 and 0.062. Their optimization budgets comprise 1,763,200 and 1,033,600 row exposures. The experiment gives all architectures the same maximum training opportunity and held-out selector. Model-specific losses, pretrained encoders, language inputs, and spatial supervision define the complete systems being compared.

The selected models supply both the offline and physical comparisons. EBV-CAN's spatial, instruction, and input-intervention diagnostics also use its selected checkpoint and training-only alternate-target templates.

The completed physical comparison uses these selected checkpoints. Main-paper Table~V pairs each policy's success counts with its matched-task offline MAE. The grounding and inference interventions use the same selected EBV-CAN checkpoint. The case analysis below documents physical behavior of the evaluated deployments.

\section{Representative Failure Cases and Recovery Priorities}
\begin{figure*}[p]
\centering\includegraphics[width=\textwidth]{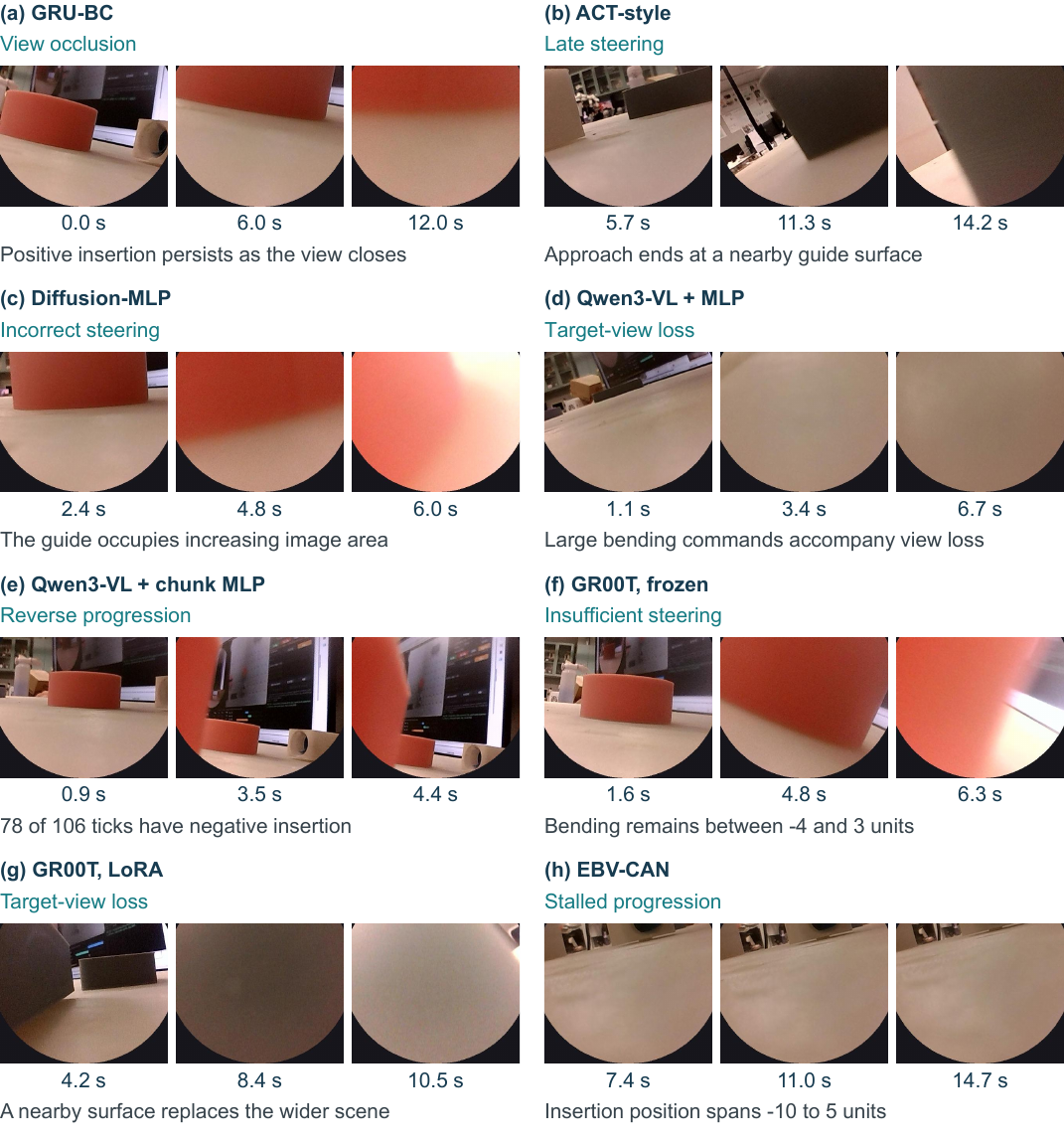}
\caption{Original EV sequences from selected L1 failures of all eight policies. Frame times are relative to each case's first recorded timestamp. Identical margins are cropped from each image to emphasize task geometry; the source images remain unchanged. Behavioral labels follow the recorded notes and aligned command evidence.}
\label{fig:failure-models}
\end{figure*}

\subsection{Evidence for Distinct Failure Behaviors}
Figure~\ref{fig:failure-models} relates local visual changes to representative failure behaviors for every policy.

\paragraph{GRU-BC and ACT-style}
In the GRU-BC occlusion example, the red guide expands to occupy most of the EV while insertion remains forward throughout execution. The insertion position increases by 48 controller units. The sequence connects continued approach to persistent view occlusion. ACT-style examples show target pass-by and incorrect, early, or late steering. The illustrated late-steering sequence moves from an open scene to a nearby guide surface, indicating that the steering response arrives after the useful approach configuration.

\paragraph{Diffusion-MLP and Qwen heads}
The illustrated Diffusion-MLP execution continues insertion as a guide fills the EV; its examples show incorrect or premature steering. The Qwen3-VL single-action head produces a median bending command of 149 against the 150-unit scale in the illustrated sequence, followed by board-dominated views. The chunk-head example applies negative insertion during 73.6\% of recorded commands and produces a net insertion-position change of $-20$ units, consistent with reverse progression. These observations distinguish steering direction, steering magnitude, and insertion direction as separate sources of failure.

\paragraph{GR00T variants}
The frozen GR00T example applies bending commands between $-4$ and 3 units while the insertion position advances by 49 units. A guide increasingly occupies the EV, consistent with insufficient steering during forward approach. Other examples exhibit late steering, target pass-by, and target-view loss. The LoRA example transitions from a broader scene to a close surface view during a large bending excursion; the bending position spans $-478$ to 8 units. Its additional failures involve incorrect, early, or insufficient steering. Backbone adaptation therefore leaves steering timing and visibility as physical control challenges.

\paragraph{EBV-CAN}
The stalled-progression example repeatedly returns to similar board-dominated views over 22.05~s. Its insertion position spans $-10$ to 5 units with a net change of 2 units. Other EBV-CAN failures show incorrect or late steering and target-view loss. Actuator movement accompanied by limited visual progress identifies the need to recognize ineffective motion and switch to a recovery behavior.

\subsection{Interpretation and Future Experiments}
The command traces connect actuator behavior to changing local views. Insertion and bending positions describe motor state, while the image sequence supplies evidence of visibility and route progress. Their joint interpretation identifies concrete recovery targets for each deployment.

Future experiments will collect corrective demonstrations around guide approach, view occlusion, target pass-by, and stalled progression. Transition-centered objectives will emphasize the timing of steering onset and release. Recovery policies will retreat and reacquire informative views after occlusion. Synchronized force and shaft-shape measurements will test whether a missed transition follows poor support formation or state-estimation error. Paired target-instruction trials and deformable three-dimensional routes will extend the present geometric and semantic evaluation.

\clearpage
\begingroup
\IEEEtriggeratref{6}
\newcommand{\BIBdecl}{\sloppy\emergencystretch=3em\microtypesetup{expansion=false}}
\putbib[references_si]
\endgroup

\end{bibunit}
\end{document}